\ifdefined\XeTeXversion\else\pdfoutput=1\fi
\documentclass[10pt]{article}
\usepackage[margin=1in]{geometry}
\usepackage{titlesec}
\titlespacing*{\section}{0pt}{1.2ex plus .3ex}{0.8ex}
\titlespacing*{\subsection}{0pt}{1ex plus .2ex}{0.5ex}
\titlespacing*{\paragraph}{0pt}{0.8ex plus .2ex}{0.6em}
\usepackage[T1]{fontenc}
\usepackage{lmodern}
\usepackage{microtype}
\usepackage{amsmath,amssymb,amsfonts,bm}
\usepackage{graphicx}
\usepackage{booktabs,multirow,array}
\usepackage{xcolor}
\usepackage{algorithm}
\usepackage{algpseudocode}
\usepackage{caption,subcaption}
\usepackage{enumitem}
\usepackage[numbers,sort&compress]{natbib}
\usepackage[colorlinks=true,linkcolor=blue!50!black,citecolor=blue!50!black,urlcolor=blue!50!black,
  pdftitle={Fix How It Acts, Not How It Thinks: Surgical Fine-Tuning of Frontier LLMs},
  pdfauthor={Sandeep Singh Sandha, Prann Bansal, Arvind Sundararajan}]{hyperref}
\usepackage{cleveref}

\graphicspath{{figures/}}
\newcommand{\Wp}{W'}
\newcommand{\Q}{Q_{\mathrm{fmt}}}
\newcommand{\sg}{\operatorname{sg}}
\newcommand{\pit}{\pi_\theta}
\newcommand{\piref}{\pi_{\mathrm{ref}}}
\newcommand{\A}{\mathcal{A}}
\newcommand{\T}{\mathcal{T}}
\newcommand{\kthree}{k_3}
\newcommand{\smaug}{Smaug-Nemotron-3.5}
\newcommand{\smaugfp}{Smaug-Nemotron-3.5-NVFP4}
\newcommand{\fig}[2][width=\linewidth]{\includegraphics[#1]{#2}}
\crefname{algorithm}{Algorithm}{Algorithms}
\Crefname{algorithm}{Algorithm}{Algorithms}

\title{\textbf{Fix How It Acts, Not How It Thinks:}\\\textbf{Surgical Fine-Tuning of Frontier LLMs}}
\author{
  Sandeep Singh Sandha \qquad Prann Bansal \qquad Arvind Sundararajan \\[4pt]
  \textbf{Abacus.AI} \\[2pt]
  \texttt{\{sandeep, prann, arvind\}@abacus.ai}
}
\date{}

\begin{document}
\maketitle

\begin{abstract}
Open-weight LLMs are competent software engineers, but they sometimes fail at tasks they are otherwise able to do.
Often it is not a failure of reasoning.
It is a few small missteps over a long trajectory (repeating a step that has already failed, or declaring the task done while a requirement is still unmet) that make the difference.
Correcting this calls for a surgical weight update that steers the model toward the right actions without interfering with its reasoning.
Such updates, typically learned with LoRA adapters, are often too small to survive merging into a released checkpoint.
In our BF16 release, the median weight change is only $1.5\%$ of a single rounding step, and merging loses roughly two thirds of the update's norm.
When the same adapter is merged into the mixed FP8/NVFP4 release of the same model, $93\%$ of the update disappears.
Instead of increasing the size of the update, we directly train on the surviving part.
We introduce \emph{merge-aware fine-tuning}.
Standard LoRA rounds the merged weights to the storage format only at export, so the loss never sees the rounded weights.
We instead round the sum of base weights and adapter update to the release format at every training step, keeping the vendor's quantization scales, and compute the loss on these weights.
The forward pass matches the serving engine numerics, and the gradient passes straight through the rounding, so the model we train is the model we ship.
The loss optimizes only the model's actions and keeps its reasoning close to the base model.
Applied to Nemotron-3.5, the recipe improves agentic coding in both the BF16 and the NVFP4 release, by up to $8.4$ points on Terminal-Bench 2.1, and we publicly release both models.
\end{abstract}

\section{Introduction}
\label{sec:intro}

Large open-weight reasoning models are becoming capable of solving real-world engineering tasks by themselves, over many turns.
However, benchmarks in software engineering demonstrate that failure modes of these models often lie in their actions, not in their reasoning.
Even a model that reasons well may still fail if a few small mistakes occur over the course of solving a long task.
This may take the form of a tool call or code patch of the wrong format~\citep{jimenez2024swebench}, repeating a step that has already failed, or declaring the task complete while a requirement is still unmet~\citep{merrill2026terminalbench}.
These are failure modes in how a model acts, not in how it thinks.

Fixing these mistakes should not require changing what the model already knows.
Fine-tuning can affect capabilities that are already working well~\citep{luo2023forgetting}, while small low-rank updates are known to result in less forgetting~\citep{biderman2024loralearnsless}.
Thus, we require a small, surgical update that improves behavior without affecting reasoning, knowledge, or other capabilities.
A \emph{surgical} update is small and targeted at a single behavior, with bounded modifications everywhere else.
Larger updates typically have more widespread effects on the model that cannot be easily predicted or verified.
Indeed, we saw this in our own runs, where checkpoints that were further from the base model had worse scores on the benchmark.
Therefore, three conditions must hold for a surgical update.
Firstly, the correct tokens need to be targeted.
Secondly, the model should move only as far as needed.
Lastly, this update must survive in the weights of the final release.

\begin{figure}[t]
  \centering
  \fig{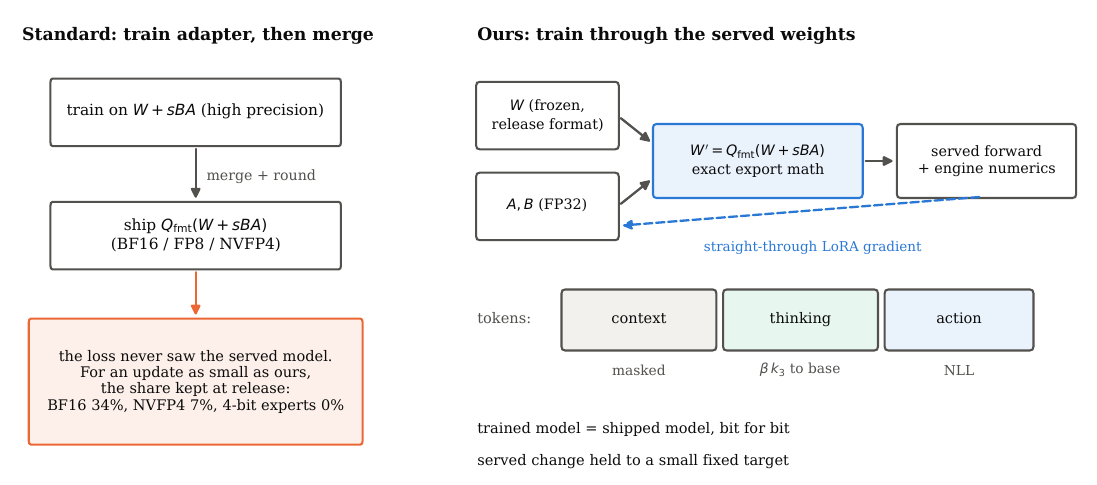}
  \caption{\textbf{Merge-aware fine-tuning.} \emph{Left:} standard LoRA trains $W + sBA$ and ships $\Q(W+sBA)$, so most of a small update is
  rounded away (\cref{sec:background}). \emph{Right:} we compute the loss through the served weight $\Wp = \Q(W+sBA)$ and pass the ordinary
  LoRA gradient straight through the rounding. Only action tokens receive a likelihood term. Thinking tokens are held near the frozen base
  by a sequence-level KL-style ($\kthree$) penalty, and context is masked. The shares kept on the left are for our released BF16 adapter
merged by plain rounding (\cref{tab:retained}). The dashed arrow is the gradient, which passes the rounding unchanged. Bottom right: the
released weights are exactly those the loss was computed on, and the served change is held to a small fixed target
(\cref{sec:method}).}
  \label{fig:overview}
\end{figure}

Low rank adaptation~\citep{hu2022lora} is the natural approach here.
The typical steps are: train the adapter weights at high precision, merge them into the base weights, and store the merged weights in the format of the base checkpoint.
The standard LoRA approach works well for the first two conditions of a surgical update.
But the last merge step is critical for the small update to survive in the release.
The merge is not just a packaging step, but actually introduces lossy data-independent rounding that can wipe out most of the update.
In our released BF16 model, a LoRA fine-tune of NVIDIA's 120B Nemotron-3.5, the median magnitude of an element in the update is only $1.5\%$ of the magnitude of one rounding step on the BF16 grid.
Roughly two thirds of the norm of the update is lost in the merge process and the surviving update only has a cosine similarity of about $0.26$ to the direction of the update that was learned during training.

This gets worse for the quantized releases.
Vendors now also offer official checkpoints in low precision~\citep{nvidia2026nemotron35supervl}, alongside BF16 models, because low-precision checkpoints are much cheaper to serve.
The 4-bit NVFP4 weights only require a fraction of the memory of BF16 weights and can run on the 4-bit tensor cores of recent GPUs~\citep{nvidia2025nvfp4pretraining}.
This allows for a smaller, faster checkpoint, which can be deployed on fewer GPUs and hence serve more users at a lower cost.
So the quantized release is typically the one deployed, and is the one a fine-tune should improve.
But the coarser quantization grid means that it is even harder for small updates to survive during merging.
When the same adapter is merged into NVIDIA's mixed FP8/NVFP4 release of the same model~\citep{nvidia2026nemotron35supervl}, $93\%$ of the update is lost, and the shared-expert weights, which are stored in 4 bits, do not change at all.

Note that this loss of the update in the merge, which we call the ``vanishing delta'' (\cref{sec:background}), is not something specific to our model.
We see it elsewhere in the literature, e.g.\ in BF16 training~\citep{zamirai2020bf16,singh2026vlafp32}, in attempts to purge knowledge from a model~\citep{zhang2025unlearningquant}, and when serving diffusion models~\citep{sglang2026loraunmerged,maxdiffusion2026loradelta}.
In our own experiments we see the same phenomenon.
We trained a normal 120B LoRA adapter on the same dataset and with the same objective we used for our release, but without the merge-aware training approach.
It loses most of the gains on the benchmarks after merging.
Interestingly, the merge keeps the same share of the update, and the same direction, for this adapter as for our release.
The difference is that our release was trained on the portion of the update that survives, and the plain adapter was not.
Theoretically, if the update were larger, more of it would survive the merge.
But this contradicts the whole idea of a small and targeted update.
Instead of scaling the update, we directly learn the part of it that survives.

There have been a number of works looking at combinations of quantization and adapters. Some work, like QLoRA-style approaches, keeps
the adapter fully separate from the quantized base model without merging~\citep{dettmers2023qlora,li2024loftq,qin2024irqlora}, while
others instead fine-tune the quantization parameters~\citep{kim2023peqa}. The closest to our setting are quantization-aware
low-rank methods which explicitly take the merged model into account during
training~\citep{xu2024qalora,jeon2025l4q,bondarenko2024lrqat,chen2025lotaqaf,guo2025intlora}, recently also for 4-bit microscaling base
models~\citep{li2026scaleqlora,axolotl2026nvfp4mergeaware,axolotl2026nvfp4native}. But here the user controls the quantization of the
target model, into a single integer or microscaling format, and typically learns or refits the scales and/or zero-points. Our setting
has three additional constraints: first, we start from a vendor's mixed BF16/FP8/NVFP4 checkpoint and we can only change the adapted
weights, and every other byte of it must stay identical; second, there are differences in numerical behavior between our training library and
the serving engine which matter at this scale; and finally, the updates are so small that almost all single weights move by less than half
a rounding step. \Cref{sec:related} discusses related work in detail.

We propose \textbf{merge-aware fine-tuning}. A release never stores the high-precision sum $W + sBA$.
It stores that sum rounded into each tensor's number format, BF16, FP8 or NVFP4.
This rounding grid is determined by the format and the vendor's scales. We
denote this stored weight as $\Wp = \Q(W + sBA)$.
We round not only at export time but at each training step, keeping the grid and scales fixed. We then train the model directly on $\Wp$, and the LoRA gradient passes straight
through the rounding to the adapter factors~\citep{bengio2013ste}.
Part of the update is still lost at the rounding, but the difference is
that training sees this loss.
The optimizer now learns an update whose surviving part leads to the desired change in the model's
behavior, not one that works only before rounding. Similarly, the numerics of the rest of the forward pass match the serving engine.
At initialization the model matches the served base weights bit for bit and each checkpoint that we release contains exactly the $\Wp$
used for the loss. In addition, we use an objective that operates on the model's behavior. Each token is classified as context,
thinking or action, and only the action tokens, namely the end of reasoning, the tool call and the end of turn, are given a likelihood
loss. Thinking tokens are not given a likelihood loss, but are instead kept close to the frozen base model by a sequence-level KL-style
drift penalty in the $\kthree$ form~\citep{schulman2020klapprox,shao2024deepseekmath}. At the same time we keep the served weights
close to the base, with a fixed target for how far they may move relative to their size. We scale learning rates
based on storage format to ensure that updates can still move weights on the coarser rounding grids.
\Cref{fig:overview} summarizes the method, and compares the two pipelines. On the left, standard LoRA trains on the high-precision sum
$W + sBA$, only rounding at export. The rounded model never appears in the loss. The change that survives the merge depends on the
size of the update compared to the rounding grid, not on the desired change in the model's behavior. For a small update like ours,
only $34\%$ survives in BF16 and $7\%$ in NVIDIA's NVFP4 release, with no change at all in its 4-bit experts (\cref{tab:retained}). On
the right, at each step of training we first construct the weights that will be stored in the release and run the forward pass using
the same numerics as the serving engine. The LoRA gradient then passes straight through the rounding unchanged to the FP32 factors $A$ and $B$, as
indicated by the dashed arrow. The row of token types indicates where we apply the loss. The last two lines summarize: the model we
train is the model we ship, and we control how much we move the served weights.

Our contributions are:
\begin{itemize}[leftmargin=*,itemsep=1pt,topsep=2pt]
  \item \textbf{Merge-aware fine-tuning for real releases.} We extend training through the merged weights to mixed BF16/FP8/NVFP4
        vendor releases, with exact quantizers that keep the vendor's scales frozen, serving-engine numerics (FP8 KV cache, the Mamba
        time-step clamp, W8A8 and W4A4 kernels), and format-aware learning rates so that coarse FP8 grids still learn (\cref{sec:method,sec:setup}).
  \item \textbf{A recipe for surgical behavior changes.} We train only on action tokens, hold the model's reasoning near the base with a
        sequence-level drift penalty, and keep the change to the served weights near a small target that applies across storage
        formats. The resulting updates change how the model acts without changing how it thinks: the model reasons about as much per step as its base,
        agentic benchmarks improve, and general capabilities are preserved (\cref{sec:method,sec:results,sec:analysis}).
  \item \textbf{A controlled comparison.} Standard LoRA with the release's data, order, learning rate and objective keeps the
        same share of its update as the release, in size and direction, but loses most of its benchmark gain once served (\cref{sec:history}).
  \item \textbf{Public releases.} We released \smaug{} and \smaugfp{} with weights and
        evaluations~\citep{abacus2026smaugnemotron,abacus2026smaugnemotronfp4}. To our knowledge, \smaugfp{} is the first publicly released fine-tune trained through the exact served weights of a vendor's
        mixed-precision release; its NVFP4 base is itself produced by quantization-aware distillation, which recovers the BF16 model's accuracy rather than updating a released checkpoint~\citep{nvidia2026nemotron35supervl,xin2026qad}. Both improve agentic coding, including
        Terminal-Bench 2.1 69.1 vs.\ 60.7 for the 120B BF16 model. These gains come without a cost to general capabilities:
        reasoning, code, instruction following, long context and vision are preserved (\cref{sec:results}).
\end{itemize}

\section{Why Small Updates Vanish at Release}
\label{sec:background}

\paragraph{The release is a rounded merge.}
A LoRA adapter~\citep{hu2022lora} learns a low-rank update $D = sBA$ applied to a frozen weight $W$. In a release, this adapter is
typically merged into the base model, rather than distributed separately. As \citet{hu2022lora} proposed, this avoids any additional
latency during inference: $W + sBA$ is computed once and stored as a normal weight. This holds in exact arithmetic. In practice,
however, the merged weight needs to be represented in the same format as the base checkpoint, which is BF16 for a full-precision
release and either FP8 or NVFP4 with fixed per-tensor or per-block scales for a quantized
release~\citep{micikevicius2022fp8,rouhani2023microscaling,nvidia2025nvfp4pretraining}. The model that the user runs will
thus operate on the rounded weight
\begin{equation}
  \Wp = \Q(W + sBA),
  \label{eq:served}
\end{equation}
rather than the high-precision sum $W + sBA$, where $\Q$ rounds onto the grid of the storage format, BF16, FP8 or NVFP4, with the
vendor's scales, whereas normal fine-tuning optimizes the latter. Provided the update is large enough,
this may not matter so much, but for the extremely small updates we are interested in here, a large fraction of the update is lost.

\paragraph{How tiny is tiny.}
For our released BF16 model, we find that the relative change in served weights for the adapted modules is only
$\rho = \lVert \Wp - W \rVert_F / \lVert W \rVert_F = 3.1\times10^{-5}$ (\cref{sec:analysis}). What does this mean for a single
weight? Numbers are only represented on a discrete grid. For a given weight, the difference between two consecutive values is one
unit in the last place, or ulp. For typical BF16 numbers, this corresponds to a relative difference of around $2^{-8}$ to $2^{-7}$ of
the weight's magnitude. An update must reach the midpoint to a neighboring value to avoid being rounded away, which requires a change
of about $0.2$--$0.4\%$ of the weight's magnitude. This is in fact quite large: most updates are significantly smaller. The median
per-element magnitude of the update $|D_{ij}|$ is actually only about $1.5\%$ of one ulp, and only $2.6\%$ of elements are updated by
at least half an ulp. In other words, these are the elements that move, and all the other values round back to their initial values.
Note also that the ulp increases with the weight's magnitude, so the same update can move a smaller weight but not affect a larger
weight. It gets much worse for lower precisions: for example, for FP8 in the E4M3 format with four exponent bits and three mantissa
bits, the grid is $16\times$ coarser than BF16 for similar magnitudes. And NVFP4 is much coarser, with only seven nonzero magnitudes
and zero available within each scaled block.

\paragraph{The ``vanishing delta''.}
We refer to this loss during merging as the ``vanishing delta''. The \emph{retained fraction} is defined as
$\kappa = \lVert \Wp - W \rVert_F / \lVert sBA \rVert_F$, where norms are taken over all adapted tensors. Note that $\kappa$ is a ratio
of norms, not an average over individual elements or tensors. This fraction is reported in \cref{tab:retained} for the adapter in our
released BF16 checkpoint, both merged into the BF16 base model and, for comparison, into NVIDIA's NVFP4 release of the same model.
(The table is split into two halves, but they both use the same adapter, so the update column is identical. The variations in the
other columns are due only to storage format.) We group the modules by type, since NVIDIA's release assigns the storage format by
module type. In BF16, a vanishing delta of around two thirds of the norm occurs during the merging process. This increases to $93\%$
in the mixed FP8/NVFP4 checkpoint, and the update to the 4-bit shared experts disappears completely: not a single NVFP4 code is
changed. Each row represents a comparison between the norm of the learned update and the norm of the actual change after rounding. For
instance, in attention projections, the update of norm 0.029 actually only changes the released BF16 weights by a norm of 0.010. The
BF16 loss is consistent over all the module groups (at $63$--$68\%$), suggesting that it is a general effect of the number format and
not something specific to a given layer. Coarser formats lose even more: the FP8 groups only retain $8$--$12\%$ of the update norm,
and the 4-bit experts do not retain any. In BF16, even though only $2.6\%$ of the elements change, around a third of the norm is
retained, because these are the ones with the biggest updates. Rounding shifts them by an entire grid step.

Because this
adapter was trained merge-aware, its released model \emph{is} the $34\%$ that survives, and the loss was computed on it. Merge-aware
training does not keep more of the update; it makes training optimize what is kept. A standard LoRA run with the same data and objective loses the same share
($\kappa = 0.343$; \cref{sec:history}) without its training ever seeing it. The same merge loss has been reported for
individual adapters merged into BF16 diffusion models at inference time~\citep{sglang2026loraunmerged,maxdiffusion2026loradelta} and for
adapters requantized into 4-bit microscaling bases~\citep{li2026scaleqlora}. Here we measure it for a 120B release in each of its storage
formats.

\paragraph{Retention depends on update size.}
Smaller updates have a bigger share of their norm lost to rounding when merging. If we merge the BF16 adapter at step~5 and
step~10 (when $\lVert sBA \rVert$ is smallest, at the beginning of training), only $7\%$ and $9\%$ of the update norm survive,
respectively. As the adapter grows larger, a bigger share of the update survives: at the released checkpoint (step~100) $34\%$ of the
update norm is retained, and at the end of training (step~158; \cref{sec:analysis}) $38\%$.

However, increasing the size of the update runs counter to our intended design goal of being a minimal and targeted modification. If
the update is larger, there is a risk of altering other aspects of the model's reasoning, knowledge and style for inputs not seen in
training. Such undesired effects are also hard to predict or verify. Increased rank and duration of fine-tuning increase the amount of
forgetting of previous capabilities~\citep{biderman2024loralearnsless}. In our BF16 run, we can see this: checkpoints that moved
further away from the base than our release solved fewer Terminal-Bench tasks (\cref{app:rho}). We are rather interested in
constraining the update to be small enough such that outside of the targeted behavior the model is still close to its base, and it is
possible to check what the update actually looks like. By constraining the update to be small, we control its size, and by setting a
target for the amount of change in served weights we can measure it (\cref{sec:ratetarget}). This is the inherent challenge: we want a
surgical update that is small by construction, and at the same time small updates are the ones that rounding tends to erase.

\begin{table}[t]
\centering
\caption{\textbf{Most of a small LoRA update does not survive a rounded merge.} We take the adapter of our released BF16 checkpoint
and merge it by plain rounding, $\Wp = \Q(W + sBA)$, into the BF16 base and, for comparison, into NVIDIA's
NVFP4 release of the same model (FP8 attention and Mamba \texttt{in\_proj}, NVFP4 shared experts; vendor scales frozen). For each
module group, $\lVert sBA\rVert_F$ is the size of the update before the merge, $\lVert \Wp-W\rVert_F$ the size of the change that
lands in the shipped weights, $\kappa$ their ratio, and ``lost'' is $1-\kappa$. For scale, $\lVert W\rVert_F \approx 1{,}354$ over
these modules. The ``all'' rows are computed over all adapted modules together, so
their norms combine the group norms as a root sum of squares, not as a sum. Bold: totals over all adapted modules.}
\label{tab:retained}
\small
\begin{tabular}{llcccc}
\toprule
Release format & Module group (storage) & update & served change & retained $\kappa$ & lost \\
 & & $\lVert sBA\rVert_F$ & $\lVert \Wp-W\rVert_F$ & & \\
\midrule
\multirow{4}{*}{BF16}  & attention $q,k,v,o$ (BF16) & 0.0293 & 0.0100 & 0.340 & 66\% \\
                       & Mamba \texttt{in\_proj} (BF16) & 0.0937 & 0.0301 & 0.321 & 68\% \\
                       & shared experts (BF16) & 0.0742 & 0.0274 & 0.369 & 63\% \\
                       & \textbf{all} & 0.1231 & 0.0419 & \textbf{0.341} & \textbf{66\%} \\
\midrule
\multirow{4}{*}{NVFP4} & attention $q,k,v,o$ (FP8) & 0.0293 & 0.0036 & 0.124 & 88\% \\
                       & Mamba \texttt{in\_proj} (FP8) & 0.0937 & 0.0073 & 0.078 & 92\% \\
                       & shared experts (NVFP4) & 0.0742 & 0.0000 & 0.000 & 100\% \\
                       & \textbf{all} & 0.1231 & 0.0082 & \textbf{0.067} & \textbf{93\%} \\
\bottomrule
\end{tabular}
\end{table}

\paragraph{Why rounding is not just noise.}
Intuitively, we might think of the rounding behaving as unbiased noise, with the true update direction still contained within the
result. However, this is not the case. Whether or not an individual element moves depends on the size of its update relative to the
local grid step (which is a function of the magnitude of the base weight, not the importance of this element in the loss function). At
the released checkpoint, we find cosine similarities of $\Wp - W$ with the trained update $sBA$ of only $0.24$--$0.28$ across the BF16
module groups, and $0.11$--$0.14$ in the FP8 groups of the NVFP4 release (\cref{sec:analysis}). For a standard LoRA adapter, trained on
the same data with the same objective, we find similar cosine similarities (\cref{sec:history}). This indicates that the poor
directionality is not a result of the merge-aware training. The served change is thus not a scaled version of the trained update.
Instead, it touches far fewer weights and points in a different direction. In the standard LoRA case, the loss has never seen this
rounded model, so evaluating the unmerged adapter and evaluating the release is not evaluating the same model.

\paragraph{The ``dormant delta''.}
This effect worsens with lower-precision formats. The served weights of a tensor remain completely constant until at least one update
is large enough to trigger a change in the stored codes. Until then, the adapter makes zero contribution to that tensor in the served
model, which is simply the base. This leads to two failure modes: in high-precision adapter training with standard LoRA, the loss can
continue to improve while the released weights stay constant. In the case of training with rounded weights, keeping the same learning
rate across all formats can freeze the served model at the base for many steps. In that case, the action loss, drift penalty and
gradient signals stay at their base values, despite the underlying adapter increasing in size. We observed this failure in an NVFP4
run with the same modules as in our BF16 run, but with a uniform learning rate across all formats. At step~80, the served weights had
only moved by $\rho = 2.7\times10^{-6}$, which is about a factor of ten smaller than in our best checkpoints. The 4-bit shared experts
did not change a single code by the time we stopped the run at step~105. The reason for this lies in the per-element updates: at the
scale of movement we used for BF16, even the $99.9$th percentile of $|D_{ij}|$ is only $0.007$ of half a grid step in NVFP4.
Therefore, it would take a learning-rate multiplier of two orders of magnitude before the stored codes begin to change in this run (\cref{sec:ratetarget}).

\paragraph{Takeaway.}
Thus, for a surgical update, merging is not merely a packaging step but a lossy, data-independent projection which discards most of
the update and changes its direction. In the experiment where we trained a standard LoRA adapter on the same data with the same
objective, we saw most of the benchmark gain vanish once the model was merged and served (\cref{sec:history}). For this reason, we
treat the rounded merge as a part of the model for the rest of this paper.

\section{Method: Merge-Aware Fine-Tuning}
\label{sec:method}

Our recipe has four parts. First, we use the exact weights that will be stored in the release for the forward pass, so the loss is
computed on the served model. Second, we replicate the numerics of the serving engine for all other forward-pass operations. Together
this ensures that the model we train is the same model that users run. Third, our objective learns from action tokens and keeps the
model's thinking close to the base. Fourth, we choose the checkpoint that we release based on a target for how far the served weights
may drift from the base, and scale the learning rates based on the storage format to ensure that weights in every format can reach the
target. An overview of the training step is given in \Cref{fig:overview}, and the full algorithm can be seen in \cref{alg:mat}.

\subsection{Training through the served weights}
\label{sec:served}

\paragraph{Served-weight forward with a straight-through gradient.}
Let us assume an adapted linear layer with a frozen weight $W$, LoRA factors of rank $r$, $A \in \mathbb{R}^{r\times d_\text{in}}$ and
$B \in \mathbb{R}^{d_\text{out}\times r}$, and a scale $s$. We first form the rounded combined weight $\Wp$ as per \cref{eq:served}.
This is the weight that we store in our release. Given a batch of row vectors $x$, we define
\begin{equation}
  h = x\,\sg[\Wp]^{\top} + \big(\ell(x) - \sg[\ell(x)]\big), \qquad \ell(x) = s\,\sg[x]\,A^{\top}B^{\top},
  \label{eq:ste}
\end{equation}
where $\sg[z]$ is a stop-gradient function: it just passes $z$ when called in the forward direction and otherwise prevents any gradient
from flowing back to $z$. The first term corresponds to the regular computation of the output of the served layer via $\Wp$, except
that no gradient flows into $\Wp$. The second term just subtracts a detached copy of $\ell(x)$ from $\ell(x)$ itself, so it has zero
value in the forward pass. Thus, we obtain the exact same output as the served layer computed with the same kernel, bit for bit. However, in the backward pass, only
the undetached $\ell(x)$ produces gradients. This ensures that $A$ and $B$ receive the same gradient they would have if the rounding
step did not exist, as in ordinary LoRA, with the upstream gradient computed through the served network. This is the application of the straight-through estimator~\citep{bengio2013ste} to $\Q$, written in the usual detach form of quantization-aware training~\citep{esser2020lsq}. The
$\sg[x]$ term makes sure that the LoRA branch does not contribute a gradient to the input as well. All gradients to $x$ then pass
through $\Wp$ alone, and are therefore computed with the weights that are actually used in the served model.

\paragraph{Dynamics of the served weights.}
As training proceeds, the high-precision factors $A$ and $B$ continue to receive tiny updates. But each element of the served weight
remains unchanged until its accumulated update reaches the rounding threshold, half a step on the local quantization grid. At that
point the element is advanced to the new grid point, and the loss observes the new weight. In other words, the update that training
improves is the rounded one that is actually served, not the high-precision update held in $A$ and $B$ before rounding.

\paragraph{Exact quantizers with frozen scales.}
$\Q$ performs the same quantization as the release export. It rounds the merged weight onto the grid of each tensor's storage format
and keeps every scale the vendor shipped. We make no attempt to refit any scales. The result is a merged checkpoint that differs from
the vendor release only in the weight bytes of the adapted modules. Details of the arithmetic and byte-level validation are in
\cref{app:validation}.

\paragraph{Exactness at initialization.}
At initialization, $B = 0$, so the update $sBA$ is zero. The base weights already lie on their own quantization grid, so
$\Wp = \Q(W) = W$. Thus the policy is identical bit for bit to the reference, and all of the drift terms of the objective (\cref{sec:objective}) are exactly zero.

\subsection{Serving-engine numerics}
\label{sec:numerics}
It is not sufficient to match the served weights in order to match the served model. Indeed, there is some numerical freedom when
implementing the inference engine, which may differ from the training library. It may use a different precision when storing
activations, use a different clamping bound, or round operations in a different order. Consequently, matching the served weights
(\cref{sec:served}) only covers part of the ``train what you ship'' requirement, and the training forward pass also needs to
numerically match the serving engine where they differ.

Whatever the serving engine is, we follow the same process: we compare the training and serving forward passes of the base model and
extract the operations for which the numerical behavior does not match. For each discrepancy, we then add to the training forward pass
a specific switch which simulates this difference, with a straight-through gradient when a rounding is involved. These switches are
checked against the corresponding engine kernels on real weights, element by element where the kernel allows it, and we also verify that the policy at initialization,
with the adapter at zero, has the served base weights bit for bit. For our deployment, we used vLLM~\citep{kwon2023vllm} with an FP8 KV
cache, according to NVIDIA's recipe for these models~\citep{nvidia2026nemotron35report}. For the BF16 release, the two numerical discrepancies were the rounding of
queries, keys and values in FP8 at the interface with the cache, and the lower bound that the training library imposes on the Mamba time step, which the serving kernel does not apply. For the quantized NVFP4 release, we additionally matched the base model's FP8 and NVFP4 kernels, and the operation which combines
the routed-expert outputs. The complete list can be found in \cref{app:validation}.

Why is this fine-grained comparison important? Because our update is by construction small: the change brought by the adapter on the
per-token action loss is a few $10^{-4}$ nats, while a single unmatched operation can change the loss by more than that. Without the
numerical matching, training would then use its limited update budget to compensate for the numerical differences between the
training library and the serving engine instead of learning the desired behavior (\cref{sec:servingnumerics}).

The point is that the merged weights depend on the storage format, while the numerical switches depend on the deployment. Any engine
which loads the release exactly and uses the same stored weights does not necessarily compute the forward pass in the same way. If the
deployment evolves, the numerical checks must be redone, and if all the switches still match then the trained model is still the
served model. Otherwise, if some operation has been modified, this equivalence does not hold anymore, and the loss was computed for a
numerically different model. The corresponding switch must be modified, and the adapter retrained.

\subsection{Learning from actions, anchoring thinking}
\label{sec:objective}

\paragraph{Token regions.}
Tokens in the training set are assigned to one of three regions, which dictates how the loss for them is computed. Context includes
all inputs to the model: system and user prompts, tool outputs, and previous turns. Thinking is the reasoning block of a supervised assistant
turn. Action includes all points of commitment in an assistant turn: the end of its reasoning, a complete tool call, and the
end-of-turn token. We supervise many assistant turns throughout each trajectory, not just the last one, while the remaining turns stay as context (\cref{sec:data}). See \Cref{fig:example} for an
example row with these labels.

\paragraph{Objective.}
Action tokens get a likelihood loss, their negative log-likelihood (NLL). Thinking tokens get none; they are held near the base by a drift penalty instead. Let $\A$ be
the action tokens and $\T$ the thinking tokens of one optimizer step, and $\T_i$ the thinking tokens of trajectory $i$. We minimize
\begin{equation}
  \mathcal{L}(\theta) = -\frac{1}{|\A|}\sum_{t\in\A}\log\pit(y_t\mid y_{<t})
  \;+\; \beta\,\frac{1}{|\T|}\sum_i |\T_i|\,\kthree(R_i),
  \label{eq:loss}
\end{equation}
\begin{equation}
  R_i = \frac{1}{|\T_i|}\sum_{t\in\T_i}\Big[\log\piref(y_t\mid y_{<t}) - \log\pit(y_t\mid y_{<t})\Big],
  \qquad \kthree(R) = e^{R} - R - 1 \;\geq 0 .
  \label{eq:k3}
\end{equation}
Here $\pit$ is the network run through the served weights (\cref{eq:ste}) and $\piref$ is the same network with the adapter
switched off, which is exactly the base. Both terms are averages over the whole optimizer step, the first per action token and the
second per thinking token. $R_i$ is the mean log-ratio over one trajectory's thinking tokens: how far the model's reasoning has
moved from the base on that trajectory. $\kthree$ turns it into a penalty that is never negative and grows like $R^2/2$ near zero.
It is the $\kthree$ function of the KL estimator of \citet{schulman2020klapprox}, written here on the log-ratio $R$ rather than on the probability ratio $e^{R}$, which GRPO~\citep{shao2024deepseekmath} applies
per sampled token; we apply it to the trajectory mean over fixed data tokens, so we call it a drift penalty rather than a KL
estimate. The two terms pull in different directions: the first makes the trajectory's actions more likely, the second pulls each
trajectory's reasoning back toward the base when its mean log-ratio has moved, in either direction. $\beta$ sets the exchange rate
between them and reads as nats of action loss per unit of $\kthree$ penalty.

\paragraph{Why the mean over a trajectory.}
We use the penalty on the mean log-ratio over the trajectory instead of one penalty for each token. In a hybrid Mamba/MoE model, even
in a single optimizer step, the routing and the rounding decisions can change, and this can displace the individual thinking-token log
probabilities by tenths of a nat without any real drift in behavior. A penalty for each token would turn those oscillations into
gradient. Taking the mean over the trajectory gets rid of this noise and allows the penalty to focus on the net drift (comparison in
\cref{sec:analysis}), but it comes at a cost: changes at each token that get compensated within the trajectory are not penalized.
Hence, the mean is a constraint on the net drift but not on the spread of per-token changes.

\begin{algorithm}[t]
\caption{\textbf{Merge-aware fine-tuning}: training, checkpoint selection and export}
\label{alg:mat}
\begin{algorithmic}[1]
\Require frozen base $\{W\}$ in release format; LoRA $\{A,B\}$ (FP32, $B_0=0$) and scale $s$; rows with region labels, in a fixed order; $\beta$; peak rate $\eta$
with warmup and cosine schedule $\eta_{k}$, format multipliers $\{\gamma_g\}$, target $\rho^\star$
\For{each step $k = 1,\dots,S$ (one batch of rows; a single pass over the data)}
\State $\Wp \gets \Q(W + sBA)$ for each adapted layer \Comment{exact export arithmetic, frozen scales}\label{alg:l:served}
\State $\log\pit \gets$ forward with $\Wp$, STE branch (\cref{eq:ste}), serving numerics\label{alg:l:fwd}
\State $\log\piref \gets$ same forward with adapters disabled, no grad\label{alg:l:ref} \Comment{$=\log\pit$ bit for bit at $k=1$}
\State $\mathcal{L} \gets$ action NLL $+\ \beta\cdot$ sequence-mean $\kthree$ on thinking (\cref{eq:loss}), step-level normalization\label{alg:l:loss}
\State clip the gradient norm; update $A,B$ with AdamW at rate $\eta_{g,k} = \gamma_g\,\eta_{k}$ per storage format\label{alg:l:update}
\State periodically: log $\rho_{k}$ $= \lVert\Wp-W\rVert_F/\lVert W\rVert_F$ over the adapted tensors and save the adapter\label{alg:l:log}
\EndFor
\State select a saved checkpoint with $\rho_{k}$ near $\rho^\star$\label{alg:l:select}
\State \textbf{export:} write its $\Wp$ bytes into the release shards; verify element-exact against a recompute of $\Q(W+sBA)$~(\cref{app:validation})\label{alg:l:export}
\end{algorithmic}
\end{algorithm}

\subsection{A target for served change and format-aware learning rates}
\label{sec:ratetarget}

\paragraph{One number for how far the model moved.}
We want the surgical update to be small, and we also want a way of measuring just how small it is. The measure we want should be a
function of the actual change in the released weights, not of the high-precision update before rounding. So we use the \emph{served
change} $\rho = \lVert \Wp - W\rVert_F / \lVert W\rVert_F$, which tells us the relative distance that the released weights have moved
away from the base weights for the adapted modules. If the weights are quantized, we dequantize them first. We take the two norms over
all the adapted tensors together, not an average of per-tensor ratios. Because $\rho$ is computed after rounding, we get only the part
of the update that survives the release. We can also think of it as $\rho = \kappa\,\lVert sBA\rVert_F/\lVert W\rVert_F$: the fraction
that makes it through, times the norm of the update relative to the base weight norm. So if the same adapter is merged into a coarser
grid, less of it makes it through and $\rho$ is smaller. For our releases $\rho$ is around $3\times10^{-5}$. In other words, the norm of
the change in released weights is only about $0.003\%$ of the norm of the base weights for the adapted modules.

\paragraph{The target.}
We keep $\rho$ around the target value of $\rho^\star \approx 3\times10^{-5}$ but do not impose it using a penalty term. Instead we
ensure a small update by a low peak learning rate, a single pass over the data, and a low-rank adapter on a small number of modules,
representing less than $0.05\%$ of the model's parameters. We log $\rho$ every five training steps and release a saved checkpoint whose
$\rho$ falls in the target range.

The target was chosen empirically based on the behavior of the model. We merge a few saved checkpoints, evaluate them on a set of
development tasks, and determine the range of $\rho$ in which the model performs best. In the case of the BF16 run, we evaluated four
checkpoints on a subset of Terminal-Bench 2.1 and found that the step-100 checkpoint (with $\rho = 3.1\times10^{-5}$) solved the most
tasks. The previous checkpoint had too small an update, and later ones had too big an update and solved fewer tasks (\cref{app:rho}).
We therefore used this range as the target. In the case of the NVFP4 run, we then selected the released checkpoint by $\rho$ alone,
before any evaluation. The range is not necessarily the same for all models, and another model might need a different range to perform
a surgical update.

\paragraph{Format-aware learning rates.}
In the merging operation, more of the update is lost when using a coarser grid. Therefore, in a case where the learning rate is
identical across the storage formats, for the same update magnitude, the FP8 weights will move less than the BF16 weights, and the
4-bit weights will not move at all (\cref{sec:background}). Thus, we set a distinct learning rate for every storage format $g$,
$\eta_g = \gamma_g\,\eta$, with $\eta$ as the peak learning rate for the run and $\gamma_g$ as a fixed scaling factor.

We determined the values of these multipliers prior to the start of the NVFP4 run, based only on the adapter from the BF16 run. We
scaled the adapter update with various factors, merged it into the FP8 grid, and determined the movement of the served weights in the
direction of the trained update. We observed that for a scaling factor of about $4$, the FP8 weights moved the same distance as the
BF16 weights had. Therefore, we set $\gamma_{\mathrm{fp8}} = 4$ and $\gamma_{\mathrm{bf16}} = 1$. A similar analysis for the NVFP4
tensors reveals that a scaling factor of two orders of magnitude is needed before the stored codes begin to change (\cref{app:nvfp4}). Scaling the
update this much would defeat our purpose of getting a small and targeted update. Therefore, we keep these tensors frozen. For the NVFP4
release, we still adapt the attention and Mamba input projections, which are stored in FP8, with the multiplier
$\gamma_{\mathrm{fp8}} = 4$, and in place of the 4-bit shared experts we adapt the MoE latent projections, which NVIDIA stores in BF16.

\Cref{alg:mat} puts the four parts together. Lines~\ref{alg:l:served}--\ref{alg:l:update} are one training step.
Lines~\ref{alg:l:log} and~\ref{alg:l:select} log $\rho$ and pick the released checkpoint by it. Line~\ref{alg:l:export} writes that
checkpoint's served weights into the release and verifies them (\cref{app:validation}).

\section{Training Data}
\label{sec:data}

\paragraph{Data sources and selection.}
We use the same data for training both 120B models: 3{,}767 rows sampled from 1{,}590 multi-turn agentic-coding trajectories,
across 489 unique tasks (\cref{tab:data}). Each trajectory is a tool-use episode produced by a stronger teacher model, in the
format expected by the respective harness. We only keep successful trajectories, where success is defined by the task. For
SWE-style tasks, all fail-to-pass and pass-to-pass tests must pass, and the submission must be clean. For terminal tasks, the
verifier must return a reward of 1. For repository tasks, the judgment must be correct. The 1{,}590 trajectories come from four
data pools, which fall into two groups.

\emph{Repository data pools.} The \textsc{lb-real} data pool is composed of 54 agentic-coding questions from
LiveBench 1.0~\citep{white2025livebench}, 16 in Python and 38 in JavaScript or TypeScript, covering nine repositories. These
questions were phased out when LiveBench 2.0 replaced them. LiveBench is developed and run by Abacus.AI. The \textsc{lb-syn} data pool
contains 293 synthetic questions we
wrote for 13 public repositories, 126 in TypeScript, 92 in Python and 75 in JavaScript. Each question is formulated as a
GitHub-style issue report on the repository, with a hidden reference fix and tests (median: four fail-to-pass tests and 1{,}372
pass-to-pass tests). We use the LiveBench 1.0 harness for grading these questions, as we do for \textsc{lb-real}, and we only keep
a question if applying the reference fix resolves it while an empty patch does not. All 969 trajectories in the \textsc{lb-syn}
data pool fix the corresponding question (median number of turns in a trajectory: 64). Neither data pool intersects with our
evaluation set. LiveBench 2.0 operates on a different codebase, a large production codebase frozen at a fixed snapshot. The skills
taught by these two repository data pools are: exploring a large and unknown codebase, understanding its structure, finding the
issue, and making a submission in the correct format.

\emph{Task-style data pools.} The \textsc{swe} data pool contains synthetic software-engineering tasks in the style of
DeepSWE~\citep{datacurve2026deepswe}. The \textsc{term} data pool contains synthetic tasks in the style of
Terminal-Bench~\citep{merrill2026terminalbench}, executed with the Terminus-2 agent. These two data pools teach the model how
the teacher deals with complex features involving multiple steps, and also how to respect the action format and submission format
expected by the two harnesses.

\emph{Beyond the training tasks.} Our models are not trained on any benchmark question. Several benchmarks used in \cref{sec:results}
are also different from the training data, either with respect to task style and evaluation harness (e.g., NL2Repo-Bench requires
the model to generate an entire repository given the repository specification, using the OpenHands scaffold), the codebase (e.g.,
LiveBench 2.0 is scored on a codebase that none of the four training pools touch), or the submission format. We observe gains
on these benchmarks while the general capabilities of the model are preserved, which suggests that the skills learned from the four
pools of data, e.g., navigating a codebase, finding an issue and solving it, are transferable to tasks beyond those seen in training,
including tasks that differ in the evaluation harness and the submission format.

\begin{table}[t]
\centering
\caption{\textbf{Training data.} Four pools of trajectories, all of them successful, from two kinds of task. The two \textsc{lb}
pools are set in public GitHub repositories. \emph{Trajectories}: the number of complete multi-turn episodes in the pool.
\emph{Distinct tasks}: the number of different questions or tasks those episodes solve. \emph{Share of the gradient}: the pool's share
of all action tokens. Only action tokens receive a likelihood loss, so this is also the pool's share of the training signal.}
\label{tab:data}
\small
\begin{tabular}{lrrr}
\toprule
Pool & trajectories & distinct tasks & share of the gradient \\
\midrule
\textsc{lb-syn} (synthetic, public repos) & 969 & 293 & 26.8\% \\
\textsc{lb-real} (LiveBench 1.0, retired) & 225 & 54 & 5.1\% \\
\textsc{swe} (synthetic SWE-style tasks) & 188 & 33 & 51.3\% \\
\textsc{term} (synthetic terminal tasks) & 208 & 109 & 16.8\% \\
\midrule
\textbf{total}                           & \textbf{1{,}590} & \textbf{489} & 100\% \\
\bottomrule
\end{tabular}
\end{table}

\paragraph{From trajectories to rows.}
We generate training rows from a prefix of a trajectory up to an assistant turn. Rows are capped at 64k tokens, so we truncate the
earliest turns if needed. We only supervise a subset of the assistant turns in a row, but keep the rest, including all of their
reasoning, as context, analogously to the served model seeing its own previous turns when it tackles a task. This means that long
trajectories yield a few overlapping rows, which tend to be long (median length about 60k tokens and 40 turns).

\paragraph{Only action tokens carry the likelihood loss.}
Most tokens in a row do not need to be imitated. Tool outputs and the model's earlier turns make up the context (83\% of all tokens),
and the reasoning of the supervised turns makes up another 11\%. Action tokens (fewer than 6\%) are the only ones that should be
imitated. This means that a standard SFT objective would mostly supervise imitating the teacher's reasoning style, which we do not want
to change. Our objective (\cref{eq:loss}) only penalizes thinking tokens through the drift penalty term, and only includes the
likelihood loss on the action tokens, making the update focused on imitating the actions rather than the reasoning leading to them.
Actions are short (median length 79 tokens): the end of the reasoning block, one tool call and an end-of-turn token.

\paragraph{What the actions teach.}
The supervised actions are mostly shell commands, which we group into a few patterns. About 39\% of them are commands that explore the
codebase (\texttt{cat}, \texttt{grep}, \texttt{sed}, \texttt{ls}, etc.), 21\% run tests, 13\% are edits, and the rest run scripts or
query the git state. The following three habits are repeated often enough in the trajectories for the model to learn them. First,
\emph{explore before acting}: the first command is exploratory in 99\% of trajectories. Second, \emph{verify before submitting}:
overall 79\% of submissions have a preceding test run, and almost all in the repository and SWE pools (the terminal tasks rarely expose
a test runner, and so the agent typically writes its own check script instead). Third, \emph{finish with the exact protocol}: in the
repository and SWE pools, the harness submission command is always used for the submission, and the terminal tasks use the two-step
completion handshake. Each action span ends in the end-of-turn token. Therefore, the data teaches the model both \emph{what} to emit and
\emph{when to stop}, which addresses exactly the action-level errors we target: malformed submissions, missing stops and premature
hand-offs. All three habits are demonstrated in \cref{fig:example}: the agent explores before editing, tests the edit, removes its
temporary test, checks a staged file that it did not modify, and then uses the exact command for the submission.

\begin{figure}[t]
\centering
\small
\setlength{\fboxsep}{6pt}
\fbox{\begin{minipage}{0.96\linewidth}\ttfamily\footnotesize
\textcolor{gray}{[C] system: tools = \{bash\} \ \ user: add AM/PM strings to one locale of a date library; 3 earlier turns}\\[2pt]
\textcolor{teal}{[T] <think> Let's take a look at the locale file.}\\
\textcolor{blue}{[A] </think> <tool\_call> bash: cat <locale file> </tool\_call> <|im\_end|>}\\
\textcolor{gray}{[C] <tool\_response> \dots\ file contents \dots</tool\_response> \ \ \dots\ 7 more exploration turns \dots}\\
\textcolor{blue}{[A] </think> bash: sed -i '<insert meridiem function>' <locale file>}\\
\textcolor{gray}{\ \ \ \ \dots\ checks the test setup, writes and runs a temporary test, runs the locale tests and the linter \dots}\\
\textcolor{blue}{[A] </think> bash: rm <temporary test> \&\& git status -{}-short \&\& git diff}\\
\textcolor{teal}{[T] <think> git status shows a staged change to a file I haven't modified \dots}\\
\textcolor{blue}{[A] </think> bash: git diff -{}-cached <that file>}\\
\textcolor{gray}{[C] <tool\_response> \dots\ a pre-existing staged change \dots}\\
\textcolor{blue}{[A] </think> bash: <run the test suites> | tail -10}\\
\textcolor{gray}{[C] Test Suites: 27 passed \ Tests: 225 passed}\\
\textcolor{teal}{[T] <think> All 225 tests passed. \dots\ Temporary test files have been deleted. Time to submit.}\\
\textcolor{blue}{[A] </think> bash: echo COMPLETE\_TASK\_AND\_SUBMIT\_FINAL\_OUTPUT \&\& git add -A <|im\_end|>}
\end{minipage}}
\caption{\textbf{A training row, trimmed.} \textcolor{gray}{[C]} context (no loss), \textcolor{teal}{[T]} thinking (sequence-level
$\kthree$ drift penalty only), \textcolor{blue}{[A]} action (likelihood loss). The trajectory explores, edits, tests, cleans up, checks the
staged diff, re-runs the suite, and submits with the exact protocol. 18 of its 21 assistant turns are supervised; the first three stay in the context without loss. The row comes from the \textsc{lb-real} pool and has 8{,}461 tokens. File paths, names and some command arguments are replaced by placeholders in angle brackets. Tool-call wrappers and the end-of-turn token are shown only on the first and last action.}
\label{fig:example}
\end{figure}

\paragraph{Batching.}
We train for a single pass over the rows. For each step we sample a fixed number of rows from each of the four pools, proportional
to their sizes, and normalize the loss over the whole step. Thus the weight on each pool is proportional to the fraction of action
tokens contributed by rows in the pool (last column in \cref{tab:data}). We do not use a curriculum. The order of rows is deterministic
and does not change between the BF16 run, the NVFP4 run, and the plain-LoRA run from \cref{sec:history}. Therefore the same rows are
seen at each step in all three runs, and we can directly compare the per-step losses.

\section{Experimental Setup}
\label{sec:setup}

\paragraph{Models.}
We fine-tune two models from the NVIDIA Nemotron family~\citep{nvidia2025nemotronh}. Both are hybrid models with Mamba-2
layers~\citep{dao2024mamba2}, mixture-of-experts (MoE) layers~\citep{shazeer2017moe}, and some attention layers. For \textbf{\smaug},
we use the BF16 checkpoint of NVIDIA Nemotron-3.5-Super-VL-120B-A12B~\citep{nvidia2026nemotron35supervl}. This model has 88 layers
consisting of 40 Mamba-2 layers, 40 MoE layers with 512 routed experts and top-22 routing, and 8 attention layers. For
\textbf{\smaugfp}, we use the official NVFP4 release of the same model from NVIDIA~\citep{nvidia2026nemotron35supervl}. This model uses
a mixed storage format: the attention and Mamba projections are in FP8 and the experts are in NVFP4. We inherit those formats in our
release. In both cases, we only adapt the language model weights. The vision and multi-token prediction (MTP) weights are
byte-identical to the base checkpoint.

\paragraph{Adapted modules and hyperparameters.}
The full configuration is provided in \cref{tab:hparams} in \cref{app:config}. Both runs use a rank-32 LoRA adapter, the same optimizer
and schedule, and $\beta = 100$. Both runs also use only a single pass over the data. We use the same settings whenever the storage
format allows it. The runs differ in the choice of modules to adapt, in the peak learning rate, and in the learning-rate multiplier for
the FP8 modules. In the BF16 run, we adapt the attention projections, the Mamba input projection, and the shared expert projections.
However, in the NVFP4 run, the shared expert projections are stored in 4-bit format, in which updates of our size do not change the
stored weights (\cref{sec:background}). We therefore instead adapt the MoE latent projections, which are stored in BF16 by NVIDIA. For
the FP8 modules, we use the learning-rate multiplier $\gamma_{\mathrm{fp8}} = 4$ discussed in \cref{sec:ratetarget}.

\paragraph{Export and verification.}
For each checkpoint, we export $\Wp$ to the vendor's original shards, using the same arithmetic as during training. We then verify the
exported checkpoint (\cref{app:validation}): every adapted weight element equals $\Q(W + sBA)$ recomputed from the base checkpoint and
the saved adapter, and all other tensors, as well as the scales and configuration files, are byte-identical to the vendor's release.
Thus, the weights we train on are exactly the weights we ship. Both releases are public and only differ from NVIDIA's checkpoints in the
adapted weights, so anyone can examine exactly what we have changed and validate it.

\paragraph{Evaluation.}
We serve all models using vLLM with an FP8 KV cache and thinking enabled. Following NVIDIA's NeMo Gym reproducibility recipes, we use
a temperature of 1.0 and top-$p$ of 0.95. There are two exceptions. For AA-LCR we use a top-$p$ of 1.0, following NVIDIA's
recipe. For the vision benchmark we use a temperature of 0.6, as specified in NVIDIA's model card. Since NVIDIA does not provide a
public recipe for this benchmark, we evaluate it using our own harness. We
divide the benchmarks into two groups. The first covers agentic coding, which is the target of our training:
Terminal-Bench 2.1~\citep{merrill2026terminalbench} with the Terminus-2 agent, SWE-bench
Verified~\citep{jimenez2024swebench,openai2024sweverified}, SWE-bench Multilingual~\citep{yang2025swesmith},
DeepSWE~\citep{datacurve2026deepswe}, NL2Repo-Bench~\citep{ding2025nl2repo}, and the official LiveBench 2.0
leaderboard~\citep{white2025livebench}. For LiveBench, we report the NVFP4 model's scores across all seven category boards, along with
the overall score. The second group covers general capabilities that we did not train for. We use these benchmarks to check whether the
update remains surgical, improving the targeted behavior without affecting other capabilities. These are GPQA
Diamond~\citep{rein2024gpqa} for reasoning, LiveCodeBench v6~\citep{jain2025livecodebench} and SciCode~\citep{tian2024scicode} for code,
IFBench~\citep{pyatkin2025ifbench} for instruction following, AA-LCR~\citep{aa2025lcr} for long context, and V*~\citep{wu2024vstar} for
vision.

\paragraph{Runs and base scores.}
We use NVIDIA's officially reported base scores, as is common when comparing against a vendor's release. The only exception is
NL2Repo-Bench, for which NVIDIA does not report a score, so we evaluate the base model ourselves. We average the SWE-bench and
Terminal-Bench scores over 2 to 6 runs. DeepSWE and NL2Repo-Bench use a single run, following the model cards. For general
capabilities, we average scores over 2 to 5 runs, or use 8 samples per question for GPQA and IFBench. The exact run counts for each
benchmark are listed in the model cards~\citep{abacus2026smaugnemotron,abacus2026smaugnemotronfp4}. Two details of the evaluation
protocol are important when interpreting the results. Firstly, both our DeepSWE scores and the base score are obtained using NVIDIA's
offline NeMo Gym harness with the OpenCode agent. These scores are therefore not comparable to the official DeepSWE leaderboard, which
uses a different agent. Secondly, for Terminal-Bench, we follow NVIDIA's NeMo Gym recipe with the Terminus-2 agent on all 89 tasks,
with one change: we retain the agent's past reasoning in the conversation history. The model's chat template removes it by default.
Under this protocol, the BF16 base model scores 63.3 on average over three runs, compared to NVIDIA's reported score of 60.7. We use
NVIDIA's figure for the headline comparisons. For the same-harness comparisons and the trajectory statistics in
\cref{sec:history} and \cref{sec:trajectories}, we use our three base runs. In every Terminal-Bench run, each task gets one attempt, and every outcome
other than a pass is counted as a failure. We rerun a task only if the harness itself crashes. We evaluate the merged plain-LoRA model
in \cref{sec:history} using the same protocol and the same vLLM engine as the release.

\section{Results}
\label{sec:results}

\paragraph{BF16 release.}
In \cref{tab:main}, we compare the two released models to their respective base models. \smaug{} outperforms the base on all
agentic-coding benchmarks. The biggest improvement is on Terminal-Bench 2.1, with 69.1 vs.\ NVIDIA's reported 60.7. On the other
benchmarks, the improvements are 2.7 points on DeepSWE, 2.0 points on NL2Repo-Bench, and 1.2 points on both variants of SWE-bench. This
is achieved through a small shift in served weights with $\rho = 3.1\times10^{-5}$. Fewer than 3\% of the adapted weights deviate from
the base weights, and all others remain bit-identical. In contrast, training a standard LoRA on the same data with the same objective
loses most of these improvements when merged and served (\cref{sec:history}).

\paragraph{NVFP4 release.}
\smaugfp{} also outperforms the NVIDIA NVFP4 base on all agentic-coding benchmarks. Furthermore, it is byte-identical to it except in
the adapted weights. The improvements vary from 0.7 points on SWE-bench Verified to 3.0 points on DeepSWE. When evaluating on the
official LiveBench 2.0 leaderboard (\cref{tab:livebench}), it achieves an overall score of 50.3 vs.\ the 46.5 of the base. It
outperforms the base on five out of seven category boards, with the biggest improvements on Coding ($+10.9$) and Browser ($+7.6$). On
the other hand, the scores decrease on Data Analysis ($-0.6$) and Research ($-3.1$). Even though the served weights are shifted to a
similar extent as for BF16, the improvement on the standard agentic benchmarks is smaller. \Cref{sec:placement} discusses this in more
detail and shows that the 4-bit format imposes additional constraints on where the update is allowed to affect the served weights.

\paragraph{General capabilities stay the same.}
The improvement in agentic performance does not come at the expense of general abilities. For both models, no reasoning, code,
instruction-following, long-context or vision score decreases by more than 0.8 points ($\Delta$ columns of \cref{tab:main}). We only
observe drops on V* in BF16 ($-0.4$) and on IFBench in NVFP4 ($-0.8$). All others are either unchanged or move up by at most 1.8
points. As we did not train for those capabilities, we read these changes as preservation rather than improvement. That is the effect
that a surgical update should have: the model is better at the task it was trained to be better at, while its general capabilities stay
roughly the same. However, fine-tuning on code does not necessarily keep those capabilities intact. For example, in a study on
Llama-2-7B, instruction tuning on code was found to cause more forgetting on general benchmarks than any other setting considered, and
more the longer training ran~\citep{biderman2024loralearnsless}. Our update is one training pass that changes the served weights by
only about $3\times10^{-5}$ and applies no likelihood loss to thinking tokens. The agent trajectories are consistent with this
reasoning/action distinction: on Terminal-Bench, we see roughly the same number of reasoning tokens per step as with the base model, but
fewer wrong answers submitted (\cref{sec:trajectories,tab:behavior}).

\begin{table}[t]
\centering
\caption{\textbf{Main results.} Ours vs.\ the base each model started from. Base numbers are NVIDIA's official figures, except where marked $^\dagger$
(NVIDIA reports no score; base run by us). Bold: better of the pair by more than 0.5. $\Delta$: ours minus base, in points. $\rho$ is computed over each model's adapted
tensors.}
\label{tab:main}
\small
\setlength{\tabcolsep}{4.5pt}
\begin{tabular}{l ccc ccc}
\toprule
 & \multicolumn{3}{c}{\textbf{120B BF16}} & \multicolumn{3}{c}{\textbf{120B NVFP4}} \\
\cmidrule(lr){2-4}\cmidrule(lr){5-7}
Benchmark & Smaug & base & $\Delta$ & Smaug & base & $\Delta$ \\
\midrule
\multicolumn{7}{l}{\emph{Agentic coding}} \\
Terminal-Bench 2.1          & \textbf{69.1} & 60.7 & $+8.4$ & \textbf{64.0} & 62.8 & $+1.2$ \\
SWE-bench Verified          & \textbf{72.3} & 71.1 & $+1.2$ & \textbf{71.8} & 71.1 & $+0.7$ \\
SWE-bench Multilingual      & \textbf{68.9} & 67.7 & $+1.2$ & \textbf{69.0} & 67.0 & $+2.0$ \\
DeepSWE [offline]      & \textbf{27.4} & 24.7 & $+2.7$ & \textbf{21.2} & 18.2 & $+3.0$ \\
NL2Repo-Bench                & \textbf{32.2} & 30.2$^\dagger$ & $+2.0$ & \textbf{30.2} & 28.3$^\dagger$ & $+1.9$ \\
\midrule
\multicolumn{7}{l}{\emph{General capabilities}} \\
GPQA Diamond                & \textbf{90.7} & 89.5 & $+1.2$ & 90.1 & 89.9 & $+0.2$ \\
LiveCodeBench v6            & 88.7 & 88.2 & $+0.5$ & \textbf{88.2} & 87.4 & $+0.8$ \\
SciCode                     & \textbf{47.7} & 47.1 & $+0.6$ & \textbf{48.4} & 46.6 & $+1.8$ \\
IFBench                     & \textbf{72.0} & 70.9 & $+1.1$ & 70.7 & \textbf{71.5} & $-0.8$ \\
AA-LCR                      & 68.8 & 68.6 & $+0.2$ & \textbf{69.0} & 67.7 & $+1.3$ \\
V* (vision)                 & 85.2 & 85.6 & $-0.4$ & 86.5 & 86.0 & $+0.5$ \\
\midrule
served change $\rho$        & \multicolumn{3}{c}{$3.1\times10^{-5}$} & \multicolumn{3}{c}{$3.5\times10^{-5}$} \\
\bottomrule
\end{tabular}
\end{table}

\begin{table}[t]
\centering
\caption{\textbf{LiveBench 2.0, NVFP4.} All seven category boards and the overall score of the official LiveBench 2.0 leaderboard
(\url{https://livebench.ai}, release 2026-10-25). Overall is the leaderboard's weighted mean of the seven boards (Long Agentic $\times2$,
Coding and Design $\times1.5$, others $\times1$). Bold: better of the pair by more than 0.5.}
\label{tab:livebench}
\small
\begin{tabular}{lccr}
\toprule
LiveBench 2.0 board & \smaugfp{} & NVFP4 base & $\Delta$ \\
\midrule
Long Agentic   & \textbf{8.4}  & 7.3           & $+1.1$ \\
Reasoning      & \textbf{75.1} & 71.7          & $+3.4$ \\
Coding         & \textbf{72.2} & 61.3          & $+10.9$ \\
Browser        & \textbf{25.6} & 18.0          & $+7.6$ \\
Data Analysis  & 65.1          & \textbf{65.7} & $-0.6$ \\
Research       & 67.8          & \textbf{70.9} & $-3.1$ \\
Design         & \textbf{62.7} & 57.1          & $+5.6$ \\
\midrule
Overall        & \textbf{50.3} & 46.5          & $+3.8$ \\
\bottomrule
\end{tabular}
\end{table}

\section{Analysis}
\label{sec:analysis}

We organize our analysis in terms of the three conditions for a surgical update explained in \cref{sec:intro}. First, the update must
survive the merge into the released weights (\cref{sec:history}). Second, it must shift the model only as much as necessary. In
\cref{sec:target}, we demonstrate that the same target shift can be applied regardless of the storage format. Third, the update must
only modify the desired behaviors. We inspect where the weights shift in the model, and how the 4-bit format restricts the shifts
(\cref{sec:placement}), and then examine the tokens to verify that the reasoning stays in place (\cref{sec:thinking}). We then look
inside the evaluation trajectories to verify that the modifications stay in the model's actions (\cref{sec:trajectories}). Finally, we
inspect the serving numerics that the model must match during training (\cref{sec:servingnumerics}).

\subsection{Standard LoRA loses most of the benchmark gain once served}
\label{sec:history}

Does training on the served model make a difference? We set up a control experiment, where we trained a standard LoRA adapter using the
same settings as our BF16 release: the exact same rows, ordering, learning rate, schedule, seed, action-only loss and thinking penalty,
but without the merge awareness. That is, its forward pass used neither the served weights nor the serving engine's FP8 attention
numerics. We trained it to the same step as our release, and merged the resulting adapter into BF16 by plain rounding
(\cref{tab:history}). If we look at the resulting merges at the level of weights, the two are almost identical: they preserve almost
the same fraction of the update ($\kappa = 0.343$ vs.\ $0.341$), have the same cosine similarity with their respective trained updates
($0.26$ each), and move the served weights by the same amount ($\rho = 3.1\times10^{-5}$).

But the similarity between the two merges does not extend to the benchmarks. The merged standard LoRA adapter achieves a SWE-bench
Verified score of 70.5, vs.\ 71.1 for the base and 72.3 for our release. On Terminal-Bench 2.1 it gets a score of 63.5, vs.\ 69.1 for
our release and 60.7 for the base. So relative to NVIDIA's reported score for the base, it only gets 2.8 of the 8.4-point gain of our
release, and relative to the base score in our own harness (63.3, \cref{sec:setup}), it is at the same level and retains no gain at all.
Both models have the same amount of change in their served weights, but only ours was trained on the change that survives merging. This
shows in the evaluation trajectories as well: the standard LoRA adapter submits wrong answers as complete at roughly the same rate as
the base model, whereas ours does so less often (\cref{sec:trajectories}).

In both merges, about $2.6\%$ of the adapted weight values change, and in both only about $9\%$ of the intended update is preserved
along the trained direction ($\kappa$ times the cosine). The difference is not in the fraction of the update that survives, but in the ability of the training to
see the surviving update: the weights that our release is trained on are the same as the ones it stores, whereas for the standard LoRA
adapter this is not the case. After merging and serving, it does not keep much of its benchmark gain. Merge-aware training inherently
avoids this mismatch, as every checkpoint that we release holds exactly the weights that were used in the computation of the loss, and
all the gains that we report in \cref{sec:results} are measured on these weights.

\begin{table}[t]
\centering
\caption{\textbf{Standard LoRA loses most of its benchmark gain once served} (120B BF16). Plain LoRA is trained like the release (same data,
order, learning rate and objective) without the merge-aware components (served-weight forward, FP8 attention numerics), and merged at the released step (100). Base scores are NVIDIA's
reported figures. Plain LoRA and release: TB 2.1 is the mean of four runs on all 89 tasks; plain-LoRA SWE-V is the mean of two runs on all 500 tasks. $\kappa$: retained fraction; cos: $\cos(\Wp{-}W, sBA)$. ``--'': not applicable.}
\label{tab:history}
\small
\setlength{\tabcolsep}{4.5pt}
\begin{tabular}{lccccc}
\toprule
Model & TB 2.1 & SWE-V & $\kappa$ & cos & $\rho$ \\
\midrule
base (NVIDIA)                   & 60.7 & 71.1 & --   & --   & 0 \\
base, our harness (3 runs) & 63.3 & -- & -- & -- & 0 \\
plain LoRA, merged             & 63.5 & 70.5 & 0.34 & 0.26 & $3.1\times10^{-5}$ \\
merge-aware release             & 69.1 & 72.3 & 0.34 & 0.26 & $3.1\times10^{-5}$ \\
\bottomrule
\end{tabular}
\end{table}

Two important points to note here. First, no error is triggered. The merge happens, and the merged checkpoint loads and serves normally.
The minor behavioral change that the update was meant to introduce is not there, and this is only noticed when inspecting the weights,
the benchmarks or the trajectories. Second, running the LoRA adapter on the side and applying it dynamically at serving
time is not a safe option either. On the BF16 path, the adapter is applied to an already rounded base output,
$\mathrm{bf16}\big(\mathrm{bf16}(xW^{\top}) + s\,xA^{\top}B^{\top}\big)$, and in our measurements on a smaller development model of the same family, this used the adapter at only
$0.4$--$0.5\times$ its strength. In an earlier 120B adapter of 232 modules, the serving engine~\citep{kwon2023vllm} also did not apply the adapter to the 80
MoE latent projections, without triggering any error. By publishing the exact merged weights that were used in training, we avoid this
class of mismatch.

\subsection{Served change as a practical target}
\label{sec:target}

\begin{figure}[t]
  \centering
  \fig{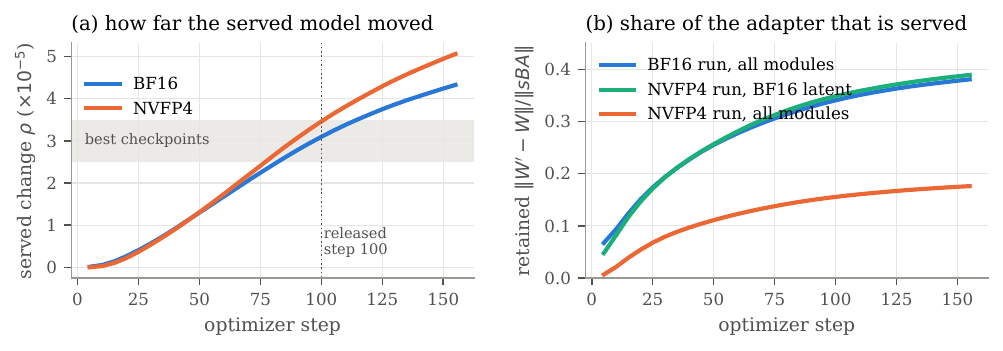}
  \caption{\textbf{Served change during training.} Both panels are measured every five optimizer steps (one step is one update on
  24 rows; 158 steps are one pass over the data) by merging the current adapter into the release format. (a) How far the served
  weights moved, $\rho$. It grows smoothly in both formats and crosses the range of our best checkpoints (shaded) near step 100,
  the step we released in both cases. (b) Share of the trained update that survives the merge, $\kappa$. BF16 weights keep
  about a third at the released step, whether in the BF16 run (blue) or in the BF16 latent projections of the NVFP4 run (green); the NVFP4
  run as a whole (orange), mostly FP8, keeps about $16\%$, so it needs a larger learning rate on its FP8 modules to reach the same
  $\rho$.}
  \label{fig:rho}
\end{figure}

\paragraph{What \cref{fig:rho} shows.}
\Cref{fig:rho} follows both releases through training. The horizontal axis is the optimizer step. Training is one pass over the data,
so moving right means a larger update. Every five steps we merge the current adapter into the release format and measure the result.
Panel (a) shows how far the served weights have moved, $\rho$. Panel (b) shows the share of the trained update that survives the merge,
the retained fraction $\kappa$. \Cref{app:rho} lists the exact values of $\rho$ for selected checkpoints.

Panel (b) makes two points. First, retention grows with the size of the update. In BF16 it rises from $7\%$ at step 5 to $34\%$ at the
released step and $38\%$ at the end of training, because a larger update pushes more values across a rounding boundary. Second,
retention is set by the storage format, not by the run. The BF16 latent projections of the NVFP4 run (green) keep the same share as the
BF16 run (blue) from about step 25 on. The NVFP4 run as a whole (orange), whose adapted weights are mostly FP8, keeps only about
$16\%$.

\paragraph{A common target across formats.}
FP8 modules keep much less of their update than BF16 modules. At the released step, the FP8 modules of the NVFP4 run keep 15 to
$18\%$ of their update, against about $34\%$ in BF16. The $4\times$ learning rate on the FP8 modules (\cref{sec:ratetarget}) compensates:
it makes the NVFP4 adapter 2.4 times larger, and the two effects roughly cancel. At the released step both runs move the served model by
almost the same amount, $\rho = 3.5\times10^{-5}$ against $3.1\times10^{-5}$ (\cref{fig:rho}a). The FP8 retention here is higher than the 8 to $12\%$
in \cref{tab:retained} because that table merges the smaller BF16 adapter into the FP8 grid, and a larger update keeps more of itself.

The shaded band in \cref{fig:rho}a is the range where our best checkpoints sat, $\rho$ between $2.5$ and $3.5\times10^{-5}$.
Checkpoints that moved further scored lower on Terminal-Bench 2.1 (\cref{tab:rho}). \Cref{sec:ratetarget} describes how this range became
the target and how it picked the NVFP4 checkpoint before any evaluation. $\rho$ is a practical guide for choosing checkpoints, not a
general predictor of quality. The right range likely depends on the data and the model.

\paragraph{Three ways to train the NVFP4 model.}
We tried three ways to train the NVFP4 release. The first adapts the same modules as the BF16 run with one learning rate for all
formats. It left training dormant. By step 80 the served weights had moved only $\rho = 2.7\times10^{-6}$, about a tenth of our best
checkpoints, and the 4-bit shared experts never flipped a single code before we stopped the run. During that time the served model was
the base, the action loss and the drift penalty sat exactly at their base values, and the adapter grew without feedback. The second
avoids coarse grids altogether: adapt only the MoE latent projections, which NVIDIA stores in BF16, at the usual learning rate. This run
trains normally but learns 0.4 times as much on held-out actions (\cref{fig:curves}a, gray). It cannot reach the Mamba input projections
or the attention weights, where the BF16 release places $57\%$ of its served change (\cref{sec:placement}). Our release takes the third
way: adapt the FP8 modules as well, with a $4\times$ learning rate. It matches the BF16 learning curve.

\subsection{Where the change lands matters}
\label{sec:placement}
Both 120B models experience similar shifts in served weights, of order $\rho \approx 3\times10^{-5}$, but their benchmark gains are not
similar. We study the location of these changes within the models via two statistics. The first is the percentage of the overall change
represented by each type of module, by squared norm. The second is the cosine similarity between the served change $\Wp - W$ and the
trained update $sBA$. This quantifies how well the retained change aligns with the direction learned during training. When the cosine is
1, they are perfectly aligned. Both are standard measures: the original LoRA study analyzes its update through the Frobenius norm
relative to the base weight~\citep{hu2022lora}, and the cosine similarity between weight changes is the usual measure of alignment in
model merging~\citep{ilharco2023taskarithmetic}.

In the BF16 model, the input projections of the Mamba layers have $52\%$ of the served change, the shared experts $43\%$ and attention
$6\%$. More than half of this change is in the last third of the model. The overall cosine similarity is $0.26$, with $0.24$--$0.28$ for
each module type.

In the NVFP4 model, the shared experts are stored in 4 bits. These do not move at all at an update of our size
(\cref{sec:ratetarget,app:nvfp4}), so the input projections make up $78\%$ of the served change, attention $12\%$ and the latent
projections $10\%$. The overall cosine similarity is $0.12$, roughly half that of the BF16 model, and is $0.11$--$0.14$ for the FP8
module types and $0.26$ for the BF16 latent projections.

These facts indicate that the reduced gains seen with the NVFP4 model are due to where the format allows a change to land, rather than
a weaker objective. Held-out action gains are very similar between the two (\cref{fig:curves}a). While similar held-out likelihood does
not necessarily imply similar downstream behavior~\citep{liu2023samepretrainingloss}, it is nonetheless clear from the original LoRA
work that not all weights are equal. Indeed, when the same number of parameters were adapted in GPT-3, selecting different weights to
adapt had a bigger impact on accuracy than adjusting the rank~\citep{hu2022lora}. In code tasks, LoRA gained most from the feed-forward
modules rather than attention~\citep{biderman2024loralearnsless}. Similarly, our NVFP4 model leaves the shared experts untouched, even
though they carry $43\%$ of the served change in the BF16 release.

\subsection{Thinking stays in place}
\label{sec:thinking}

\begin{figure}[t]
  \centering
  \fig{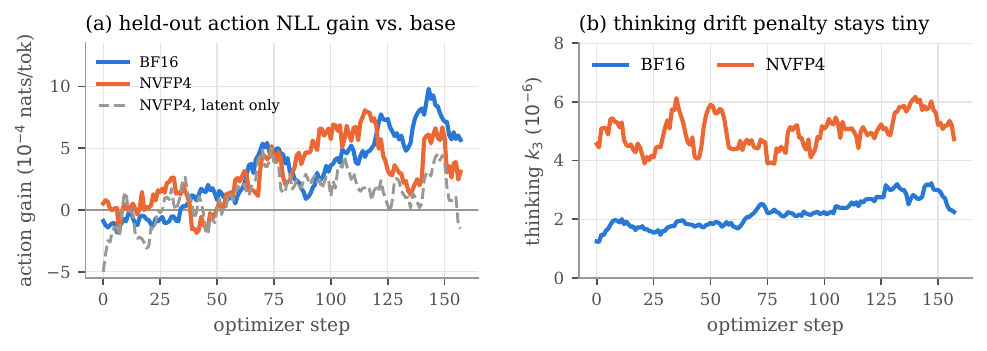}
  \caption{\textbf{Learning curves.} Training is a single pass, so each step's rows are unseen when they are scored; every point is a
  held-out measurement (10-step mean). (a) Action gain: the drop in negative log-likelihood (NLL) of the teacher's actions relative to the
  base, per action token. Gains appear after about 60 steps, once the served model has moved enough ($\rho \gtrsim 1.5\times10^{-5}$,
  \cref{fig:rho}a). NVFP4 with format-aware learning rates (orange) learns as much as BF16 (blue); an NVFP4 run that adapts only the
  BF16 latent projections (gray) learns $0.4\times$ as much. (b) The thinking drift penalty $\kthree$ against the base stays at
  1 to $6\times10^{-6}$, which corresponds to a mean log-ratio of about 0.0015 to 0.0035 nats per thinking token, since
  $\kthree(R) \approx R^2/2$: actions improve while the reasoning stays in place.}
  \label{fig:curves}
\end{figure}

\paragraph{What \cref{fig:curves} shows.}
\Cref{fig:curves} shows the evolution of the two releases as we train. Note that we make a single pass through the data, and score all
24 rows of each step before updating the model weights. As such, the rows used in scoring have not been trained on yet, and each point
is a held-out measurement. Panel (a) shows the action gain, defined as the decrease, relative to the base, in the negative
log-likelihood per action token of the teacher's actions. Panel (b) shows the thinking drift penalty $\kthree$, which measures how far
the model's thinking has drifted away from the base. The action gain remains nearly zero for the first 60 steps. In this phase, the
served weights themselves have barely moved either, with $\rho$ under about $1.5\times10^{-5}$ (\cref{fig:rho}a). The action gain
subsequently rises, averaging around $5\times10^{-4}$ nats per token over the remaining steps, and $+0.00031$ and $+0.00030$ for BF16
and NVFP4 respectively over all steps. The thinking drift penalty meanwhile remains in the range of $1\times10^{-6}$ to
$6\times10^{-6}$. Hence the update has made a positive change to the model's actions, while keeping the thinking close to the base. This
is the separation we designed the data and objective to achieve (\cref{sec:data}), and it is also borne out in the evaluation
trajectories below.

\paragraph{The drift penalty is a guard, not the driver.}
The thinking $\kthree$ term stays in the range of $1\times10^{-6}$ to $6\times10^{-6}$ throughout (\cref{fig:curves}b). The 10-step
mean is around $1$--$3\times10^{-6}$ for BF16 and $4$--$6\times10^{-6}$ for NVFP4. Hence the $\beta\cdot\kthree$ term contributes under
$0.5\%$ of the loss, and only around $0.1$--$0.2\%$ on average. The effect of the thinking drift penalty is therefore mostly that of a
guard. The main reason thinking remains close to the base is that thinking tokens are never a target of likelihood optimization and the
weight update is small. In fact, the NLL of thinking decreases a little, by around $0.001$ nats per token by the end of training,
although thinking is never trained on. So the reasoning that precedes an action becomes slightly more likely alongside the better
prediction of the actions themselves.

\paragraph{Actions move, thinking does not.}
We see this same effect when looking at the evaluation trajectories. The \smaug{} model uses as many reasoning tokens per step as the
base model does, and its actions have the same length. However, it repeats fewer steps, and is less likely to submit an incorrect answer
as complete (\cref{sec:trajectories}).

\paragraph{Sequence-mean vs. per-token penalty.}
With a per-token penalty, in development runs on a smaller model from the same family, we observed that the effects of routing
and rounding kept the penalty at a noise floor of 0.02 to 0.04 nats. The norm of the gradients also increased by about tenfold, causing gradient clipping to fire. With the sequence-mean
form used here (\cref{eq:k3}), we did not observe either of these issues. In our released runs, gradient clipping was never even
approached.

\subsection{What changes inside the trajectories}
\label{sec:trajectories}

\begin{table}[!tp]
\centering
\caption{\textbf{What changed on Terminal-Bench 2.1} (120B BF16) for the base, the standard LoRA adapter of \cref{sec:history} and our
release. \emph{Top:} how the 89 tasks ended, as mean tasks per run over three base runs and four runs per adapter. The first four rows add
up to 89 before rounding. Precision is the share of all completion claims that passed, pooled over the runs. A few tasks passed without
a claim, so it is not the ratio of the rows above. \emph{Bottom:} the two failure
modes of the Terminal-Bench 2.0 taxonomy that differ between models, as the share of each model's judged trajectories on the 58 judged
tasks. How the trajectories were chosen and judged is explained in
\cref{sec:trajectories}; definitions and all nine modes are in \cref{app:judge}.}
\label{tab:behavior}
\small
\begin{tabular}{lccc}
\toprule
 & base & plain LoRA & release \\
\midrule
\multicolumn{4}{l}{\emph{How the 89 tasks ended (mean tasks per run)}} \\
passed (higher is better)             & 56.3 & 56.5 & 61.5 \\
declared complete, scored as failed (lower is better) & 23.7 & 25.3 & 19.8 \\
ended without a completion claim (lower is better) &  1.3 &  1.0 &  0.5 \\
hit the three-hour limit (lower is better) &  7.7 &  6.3 &  7.3 \\
precision of completion claims (\%, higher is better) & 69.7 & 68.5 & 75.5 \\
reasoning tokens per call (median, unchanged is desired) & 2{,}144 & 2{,}001 & 2{,}055 \\
\midrule
\multicolumn{4}{l}{\emph{Judged failure modes (\% of each model's trajectories on the 58 judged tasks)}} \\
step repetition (lower is better)        & 45.4 & 37.6 & 36.8 \\
no failure mode (higher is better)       & 15.5 & 17.9 & 23.0 \\
\bottomrule
\end{tabular}
\end{table}

\Cref{tab:behavior} shows results for the base model, the standard LoRA adapter from \cref{sec:history} and \smaug{} on Terminal-Bench
2.1. We take the three runs of the base model on our harness (\cref{sec:setup}) and the four runs used for reporting the score of each
adapter. From the outcome rows we see that \smaug{} passes more tasks and does not declare failed tasks as complete as often. Of its
claims to task completion, 75.5\% are successful, while the base model has 69.7\%. Higher precision means fewer false claims of
success. However, on a per-step basis we do not see much
difference between the models, for example in the reasoning tokens per call (we take the mean per call for each task-run and then the
median over task-runs). To help understand what is changing throughout a task, we used an LLM judge to analyze whole trajectories using
the failure taxonomy of the Terminal-Bench 2.0 paper~\citep{merrill2026terminalbench}, which comprises nine failure modes in three
classes: execution, coherence and verification. See \cref{app:judge} for details, full tables and calibration checks.

\paragraph{Which trajectories.}
Only considering the failed trajectories would be an unfair comparison, as the release fails on a harder subset of tasks than the base
model, so its failures would be compared against easier failures from the base model. Instead, we perform an evaluation of all three
models on the same set of tasks. We use three runs of each model on the same harness (\cref{sec:setup}) and take each task which at
least one of the models failed at least once. This yields 59 of the 89 tasks. The judge declined one of them for every model, which leaves 58. We then judge all nine trajectories (both successes and
failures) for each of these tasks. This means that we are comparing all three models on exactly the same task mix, and no difference
between them can be attributed to which tasks each model failed. For failed trajectories, we tell the judge which tests failed.
For successes, we do not tell the judge the result. We check
the judge in two ways in \cref{app:judge}. First, when twenty failed trajectories are judged again with the result withheld, the judge
reproduces $93\%$ of its own mode decisions. Second, the judge still assigns modes on tasks which neither the base nor the release ever fails, so it does
not simply leave successful runs unlabeled. Each of the cells in the lower half of \cref{tab:behavior} is the percentage of that model's
judged trajectories which were labeled with the given mode. A given trajectory might have multiple of the nine modes assigned to it, or
none. Consequently, the three columns show independent percentages, and are not expected to sum to a constant.

\paragraph{What changes.}
The two rows in the lower block of \cref{tab:behavior} move with both adapters.
They move in the same direction.
For the release, the ranges separate from the base's.
For every other mode, the adapters' three-run ranges overlap the base's
(\cref{tab:app_modes}). The
first is step repetition, where lower is better: re-running a command or re-applying a fix that has already failed wastes steps and
is a sign that the agent is not adapting its approach. On these tasks the base model repeats a step in $45\%$ of its trajectories, compared to $38\%$ for the plain
adapter and $37\%$ for \smaug{}. Among passed trajectories alone, the base repeats a step in $35\%$ of its runs, the plain adapter
in $21\%$ and \smaug{} in $23\%$, and the three-run ranges of both adapters lie entirely below the base's, so the drop is not a byproduct
of passing more tasks.

The second is the share of trajectories without any of the nine modes, that is, trajectories in which the judge found nothing to
fault: no repeated step, no false completion claim, no contradiction between plan and commands, no missing or weak verification. Higher
is better. Here the three models differ, with $16\%$ for the base, $18\%$
for the plain adapter, and $23\%$ for \smaug{}. Every \smaug{} run has more trajectories without any of the nine modes than every base and
plain-LoRA run.

\paragraph{What does not change.}
Note that all other modes are within a few points of the base model. For each of them, the share in the release's three runs
overlaps the share in the base's three runs, so these differences are within the run-to-run variation of a single model
(\cref{tab:app_modes}). Step repetition and the share with no failure mode are the only modes for which the two ranges do not overlap.
Verification habits are very similar for the three models, and do
not depend on the outcome: around half of all successful trajectories are assigned a verification-mode label in each model, and this
fraction is also what we see on tasks that neither the base nor the release ever fails. These labels therefore seem to capture a habit
inherited from the base model, which neither update touches, rather than an explanation of its failures. The plain adapter shares
the drop in step repetition, but fewer of its trajectories are free of every mode, it passes fewer tasks, and its completion claims are
wrong at the base's rate (\cref{tab:behavior}).

\paragraph{Limits.}
These labels are produced by an LLM judge. We use the taxonomy and, with the adaptations listed in \cref{app:judge}, the rubrics that the Terminal-Bench authors validated against human
annotators~\citep{merrill2026terminalbench}. Because the differences in individual modes are
only a few points, we regard them as patterns that hold across runs, not as individually significant effects. We present these results
because a judge that reads the transcript can observe aspects of model behavior that a test suite does not.

\subsection{Serving numerics are part of the model}
\label{sec:servingnumerics}
As with the weights, the serving numerics are part of the model that is shipped.
They are a property of the deployment, not of the checkpoint.
One engine stores the KV cache in FP8 and another in BF16, kernels round in different orders, and clamping bounds differ between
libraries.
A small update has to be trained against the engine that will serve it, and a different deployment calls for a different set of
emulated details.
We served with NVIDIA's own recipe for these models, so that our scores are comparable with NVIDIA's reported numbers.
Against that recipe, the training library's forward pass differed in two details.
First, the engine rounds queries, keys and values to FP8 at the KV-cache boundary.
Second, it omits the lower bound that the training library enforces on the Mamba time step (\cref{sec:numerics}).
Both mismatches produce a loss at initialization that differs from the loss of the served base model.
In development runs on a smaller model from the same family, we found that omitting the FP8 rounding on the training side resulted in
more than half of the adapter's gain being lost upon serving.
Keeping the time-step clamp of the training library sent most of the gradient into compensating for the mismatch, with gradient
clipping firing on every step.
We therefore emulate both details during training, so that at initialization the model has the served base weights bit for bit and its loss is the loss of the served base, and
training can only change the model through the update it optimizes.
For an update this small, that match is necessary.

\section{Related work}
\label{sec:related}

\paragraph{Adapters, quantization and exact merges.}
LoRA~\citep{hu2022lora} and its variants like DoRA~\citep{liu2024dora} build fine-tuning updates around low-rank products. Some
work runs these adapters beside a quantized base model. QLoRA fine-tunes adapters on a frozen 4-bit base~\citep{dettmers2023qlora},
LoftQ initializes them to correct the quantization error~\citep{li2024loftq}, and IR-QLoRA retains more information in the quantized
base~\citep{qin2024irqlora}. None of these three methods merges the adapter into the quantized base. What separates these methods from merge-aware fine-tuning
is where the quantizer sits. In QLoRA it sits on the frozen base, upstream of the adapter: the forward pass dequantizes the 4-bit weights, adds the
16-bit update $sBA$ beside them, and the gradient reaches $A$ and $B$ through those dequantized weights. The update itself is never
rounded, and it is released as a separate adapter. LoftQ keeps this arrangement and moves the repair to initialization, choosing the
quantized base $\hat W$ and the adapter jointly so that their sum approximates $W$ before training. Our setting has no such error to
repair, because the base already lies on its grid, $\Q(W)=W$. The problem is the reverse one, an update too small to leave the grid. In
merge-aware fine-tuning the quantizer sits on the sum, $\Wp=\Q(W+sBA)$, downstream of the adapter, and $A$ and $B$ are trained through
that rounding with a straight-through estimator (\cref{eq:ste}). The trained model is the merged model. The closest
techniques to our setup account for the merged model during training. QA-LoRA uses group-wise pooling to constrain the adapters so that
they can be folded exactly into the zero-points of the integer quantizer~\citep{xu2024qalora}. L4Q merges the adapter into a full-precision weight before
quantization. It jointly trains the quantizer's scale and zero-point with the adapter via a straight-through estimator and deploys
the quantized merged weight~\citep{jeon2025l4q}. LR-QAT inserts a low-rank update inside an integer quantizer to make quantization-aware
training memory-efficient~\citep{bondarenko2024lrqat}. LoTA-QAF integrates ternary adapters in integer weights so that the merged weights
lie on the quantization grid~\citep{chen2025lotaqaf}. IntLoRA adapts quantized diffusion models with integer low-rank factors. These
factors are merged via integer multiplication or bit shifts, without any extra post-training quantization step~\citep{guo2025intlora}. FALQON trains directly on an FP8 merged backbone, with training speed as its goal, and applies only the rows with the largest accumulated update at each step, because small changes produce no effective update in FP8~\citep{choi2025falqon}. On
native 4-bit microscaling checkpoints, Scale-QLoRA freezes all E2M1 codes and adapts the block scales only. Here the merge is exact because only the stored scales change. It also demonstrates that a naive merge followed by a requantization step can wash away the adaptation~\citep{li2026scaleqlora}.
Our forward pass follows the L4Q construction, which Scale-QLoRA also runs as a baseline on native 4-bit grids, but we apply it to a vendor's mixed-format release with frozen instead of learned or recomputed scales. We
consider BF16 rounding, FP8 with the vendor per-tensor scales and NVFP4 with its block scales. Scale-QLoRA moves the block scales and
freezes the codes, whereas we keep every vendor scale and change only weight bytes. The reason for freezing these scales is that
it guarantees that everything except the weight bytes of the adapted modules, including their scales, is bit-identical to the vendor release.

\paragraph{Open-source implementations.}
Merge-aware LoRA is also implemented in open-source training software. The Axolotl trainer added training through $\Q(W+sBA)$ for NVFP4
bases in July and September 2026, recomputing block scales on the base's per-tensor grid and producing a merged
checkpoint which Axolotl reports as bit-exact~\citep{axolotl2026nvfp4mergeaware,axolotl2026nvfp4native}, and a BF16 variant for reinforcement learning with merged vLLM
weight sync on 27 October 2026~\citep{axolotl2026bf16mergeaware}. We apply the same construction to a vendor's mixed-precision release at
120B scale, with a behavioral objective. Our publicly released \smaug{} (BF16, released 18 October 2026) and \smaugfp{} (mixed FP8/NVFP4,
released 22 October 2026), both 120B models, were trained this way, and their weights and evaluations are
available~\citep{abacus2026smaugnemotron,abacus2026smaugnemotronfp4}. To our knowledge, \smaugfp{} is the first publicly released fine-tune trained
through the exact served weights of a vendor's mixed-precision release. Scale-QLoRA runs the same construction as a baseline on
single-format NVFP4 and MXFP4 checkpoints, including gpt-oss-120B, and does not release fine-tuned models~\citep{li2026scaleqlora}.

\paragraph{Quantization-aware training and small updates.}
Fake quantization with full-precision shadow weights~\citep{jacob2018quantization}, trained through the straight-through
estimator~\citep{bengio2013ste}, is the basis of QAT, including for LLMs~\citep{liu2024llmqat,chen2025efficientqat}. We apply the same idea with frozen base
weights and a low-rank shadow update, on a fixed vendor quantization layout. That small updates vanish in low precision is known in several
settings: for accumulation during training~\citep{gupta2015limited,zamirai2020bf16}, for BF16 weights fine-tuned without an FP32 master
copy, where most optimizer updates are rounded away~\citep{singh2026vlafp32}, for FP4 training below a gradient-to-noise
threshold~\citep{chmiel2025fp4}, and for LoRA updates applied to an FP8 backbone~\citep{choi2025falqon}. \citet{zhang2025unlearningquant} show that 4-bit post-training quantization, round-to-nearest and also
GPTQ~\citep{frantar2023gptq}, undoes small unlearning updates. We quantify the effect for fine-tuning deltas at release time, across BF16,
FP8~\citep{micikevicius2022fp8} and NVFP4~\citep{nvidia2025nvfp4pretraining}, a block-scaled format related to
microscaling~\citep{rouhani2023microscaling}, and connect it to benchmark outcomes. We train through it, rather than avoiding it with
larger learning rates or by keeping the adapter unmerged.

\paragraph{Mismatch between training and serving.}
In reinforcement learning, differences between the rollout engine and the trainer turn nominally on-policy updates into off-policy
ones~\citep{yao2025offpolicy}, and BF16 rounding has been identified as the root cause~\citep{qi2025fp16mismatch}. We face the same gap in
supervised fine-tuning. A training library computes the loss, but the model is judged by a serving engine. We close the gap by emulating
the engine's numerics operation by operation (FP8 KV cache, the Mamba time-step clamp, W8A8 and W4A4 kernels), rather than by changing the
precision on either side.

\paragraph{Model merging.}
Task arithmetic treats fine-tuning deltas as vectors that can be added and scaled~\citep{ilharco2023taskarithmetic}. TIES and DARE show
that deltas are redundant and can be trimmed or dropped~\citep{yadav2023ties,yu2024dare}. We study a complementary failure. With a single
model and no interference, the delta can still be erased by the rounding that writes it into the release format.

\paragraph{Targeted changes to a model.}
Model editing rewrites individual facts by locating the weights that store them. ROME edits one MLP layer per
fact~\citep{meng2022rome}, and MEMIT extends this to thousands of facts at once~\citep{meng2023memit}. Surgical fine-tuning, in the sense of \citet{lee2023surgical}, adapts a model to
distribution shift by tuning only a chosen subset of layers. Our goal is related but differs in what is changed and how. We change a
\emph{behavior}, how the model acts in agentic tasks, rather than a stored fact. We do it with ordinary low-rank fine-tuning on
demonstrations rather than by locating weights. And we keep the change small with a target for the served change. All of these methods
aim for a change that stays local to its target. Ours adds the requirement that the change survives the release format.

\paragraph{Preserving capabilities and reasoning.}
Fine-tuning causes forgetting~\citep{luo2023forgetting}, and LoRA trades less learning for less forgetting~\citep{biderman2024loralearnsless}.
Forgetting tracks the KL divergence from the base on the new task's distribution~\citep{shenfeld2025rlrazor}. Relatedly, likelihood training on one response shifts probability mass across many
others~\citep{ren2025learningdynamics}. KL penalties to a reference policy are standard in RLHF~\citep{ouyang2022instructgpt}. We anchor
only thinking tokens, with the non-negative $\kthree$ form~\citep{schulman2020klapprox} applied to sequence means. We keep the update small in two
places: in weight space, with a target for the served change, and in output space, with the drift penalty on thinking.

\paragraph{Agent tuning.}
SFT on successful agent trajectories is a common recipe~\citep{zeng2024agenttuning,chen2023fireact}, extended to software engineering by
SWE-Gym and SWE-smith~\citep{pan2025swegym,yang2025swesmith}. Agent-FLAN separates format following from reasoning in agent
data~\citep{chen2024agentflan}. Structured Agent Distillation segments trajectories into reasoning and action spans and supervises them
with separate masks~\citep{liu2026sad}. EKSFT masks tokens with high entropy or high KL divergence from a reference model~\citep{liu2026eksft}.
Recent work argues that supervising observation tokens as well changes how agents explore under RL~\citep{zhang2026actobs}, so action-only
supervision is not universally preferable. We use it for small updates to an already strong model, where the goal is to leave reasoning in
place. We contribute a loss split for such data, likelihood on actions and a
drift penalty on thinking, together with a target for the served change and training in the served format.

\section{Limitations}
\label{sec:limitations}

\paragraph{Evaluation.}
We omit the full benchmark suite for the checkpoints we did not release, as evaluating 120B agentic models consumes the majority of
our compute budget. One run of Terminal-Bench 2.1 takes about five hours on two B300 GPUs.
Consequently, we chose to spend the evaluation budget on the released models instead of running the full suite for each ablation.

\paragraph{Baselines.}
We compare against standard LoRA (\cref{sec:history}) but not against other methods for retaining small updates during merging.
Stochastic rounding~\citep{maxdiffusion2026loradelta} retains the update in expectation, but the resulting rounded model is not the
same as the model the loss was computed for. Alternatively, one could fine-tune standard LoRA with a higher learning rate to retain a
larger share of the update during merging. However, this would also increase the change to the served weights, which contradicts the
goal of having a small, targeted update. We found in our own run that checkpoints which drifted more from the base model than our
release had worse scores (\cref{app:rho}), which is why we did not explore this direction.

\paragraph{Scope.}
The loss of the update during merging that we see here is specific to small updates and coarse storage formats. Merge-aware training is thus most relevant when the update is intentionally made
small, or when the released model is quantized.

\section{Conclusion}
\label{sec:conclusion}
A surgical update to a strong model should be small and targeted, but such updates often fail to survive the standard model release
pipeline. Training a LoRA adapter and merging it into the checkpoint can lead to rounding that drops most of the update. The intended change in behavior can get lost along with the update. Merging is thus not just a packaging
step, but part of the model that needs to be taken into account during training. A surgical update needs to target the right tokens,
make just the necessary changes to the model and still be there in the released weights. Standard LoRA does not guarantee the last
condition. Merge-aware fine-tuning achieves this by using the exact weights that will be released and the same numerics as the serving
engine to compute the loss. We are thus training the model we ship. It does not mean that we succeed in merging more of the update, but
instead that we directly train the part that will survive. We combine this with a likelihood loss on the action tokens only, a
sequence-level penalty to keep thinking close to the base model, and a target that bounds the change in served weights. This was
applied to a 120B model, leading to improvements in agentic coding for both BF16 and NVFP4 with preserved general capabilities. We
expect the same principle, train what you ship, to be useful in any place where small updates need to survive rounding into coarse
number formats, such as unlearning, safety patches, personalization and on-device models.

\bibliographystyle{plainnat}
\bibliography{refs}

\begin{thebibliography}{77}
\providecommand{\natexlab}[1]{#1}
\providecommand{\url}[1]{\texttt{#1}}
\expandafter\ifx\csname urlstyle\endcsname\relax
  \providecommand{\doi}[1]{doi: #1}\else
  \providecommand{\doi}{doi: \begingroup \urlstyle{rm}\Url}\fi

\bibitem[{Abacus.AI}(2026{\natexlab{a}})]{abacus2026smaugnemotron}
{Abacus.AI}.
\newblock {Smaug-Nemotron-3.5}.
\newblock Hugging Face model card and weights, released 18 October,
  2026{\natexlab{a}}.
\newblock URL \url{https://huggingface.co/abacusai/Smaug-Nemotron-3.5}.
\newblock Accessed 23 November 2026.

\bibitem[{Abacus.AI}(2026{\natexlab{b}})]{abacus2026smaugnemotronfp4}
{Abacus.AI}.
\newblock {Smaug-Nemotron-3.5-NVFP4}.
\newblock Hugging Face model card and weights, released 22 October,
  2026{\natexlab{b}}.
\newblock URL \url{https://huggingface.co/abacusai/Smaug-Nemotron-3.5-NVFP4}.
\newblock Accessed 23 November 2026.

\bibitem[{Artificial Analysis}(2025)]{aa2025lcr}
{Artificial Analysis}.
\newblock Announcing {AA-LCR}: {Artificial Analysis} long context reasoning
  benchmark.
\newblock \url{https://artificialanalysis.ai/articles/announcing-aa-lcr},
  August 2025.
\newblock 100 questions over documents of about 100k tokens.

\bibitem[{Axolotl contributors}(2026{\natexlab{a}})]{axolotl2026bf16mergeaware}
{Axolotl contributors}.
\newblock Add {BF16} merge-aware {LoRA} for merged {vLLM} sync and export.
\newblock GitHub pull request axolotl-ai-cloud/axolotl \#4216, merged 27
  October, 2026{\natexlab{a}}.
\newblock URL \url{https://github.com/axolotl-ai-cloud/axolotl/pull/4216}.
\newblock Accessed 23 November 2026.

\bibitem[{Axolotl
  contributors}(2026{\natexlab{b}})]{axolotl2026nvfp4mergeaware}
{Axolotl contributors}.
\newblock feat: sonicmoe nvfp4 lora merge aware.
\newblock GitHub pull request axolotl-ai-cloud/axolotl \#3822, merged 16 July,
  2026{\natexlab{b}}.
\newblock URL \url{https://github.com/axolotl-ai-cloud/axolotl/pull/3822}.
\newblock Accessed 23 November 2026.

\bibitem[{Axolotl contributors}(2026{\natexlab{c}})]{axolotl2026nvfp4native}
{Axolotl contributors}.
\newblock Support native {NVFP4} {LoRA} backends and merge-aware training.
\newblock GitHub pull request axolotl-ai-cloud/axolotl \#4045, merged 28
  September, 2026{\natexlab{c}}.
\newblock URL \url{https://github.com/axolotl-ai-cloud/axolotl/pull/4045}.
\newblock Accessed 23 November 2026.

\bibitem[Bengio et~al.(2013)Bengio, L{\'e}onard, and Courville]{bengio2013ste}
Yoshua Bengio, Nicholas L{\'e}onard, and Aaron Courville.
\newblock Estimating or propagating gradients through stochastic neurons for
  conditional computation.
\newblock arXiv preprint arXiv:1308.3432, 2013.

\bibitem[Biderman et~al.(2024)Biderman, Portes, Gonzalez~Ortiz, Paul,
  Greengard, Jennings, King, Havens, Chiley, Frankle, Blakeney, and
  Cunningham]{biderman2024loralearnsless}
Dan Biderman, Jacob Portes, Jose~Javier Gonzalez~Ortiz, Mansheej Paul, Philip
  Greengard, Connor Jennings, Daniel King, Sam Havens, Vitaliy Chiley, Jonathan
  Frankle, Cody Blakeney, and John~P. Cunningham.
\newblock {LoRA} learns less and forgets less.
\newblock \emph{Transactions on Machine Learning Research (TMLR)}, 2024.

\bibitem[Bondarenko et~al.(2024)Bondarenko, Del~Chiaro, and
  Nagel]{bondarenko2024lrqat}
Yelysei Bondarenko, Riccardo Del~Chiaro, and Markus Nagel.
\newblock Low-rank quantization-aware training for {LLMs}, 2024.
\newblock ICML 2024 Workshop on Efficient Systems for Foundation Models
  (ES-FoMo-II).

\bibitem[Chen et~al.(2023)Chen, Shu, Shareghi, Collier, Narasimhan, and
  Yao]{chen2023fireact}
Baian Chen, Chang Shu, Ehsan Shareghi, Nigel Collier, Karthik Narasimhan, and
  Shunyu Yao.
\newblock {FireAct}: Toward language agent fine-tuning.
\newblock arXiv preprint arXiv:2310.05915, 2023.

\bibitem[Chen et~al.(2025{\natexlab{a}})Chen, Li, Peng, Wang, Ren, Shi, and
  Hu]{chen2025lotaqaf}
Junyu Chen, Junzhuo Li, Zhen Peng, Wenjie Wang, Yuxiang Ren, Long Shi, and
  Xuming Hu.
\newblock {LoTA-QAF}: Lossless ternary adaptation for quantization-aware
  fine-tuning.
\newblock In \emph{Advances in Neural Information Processing Systems
  (NeurIPS)}, 2025{\natexlab{a}}.

\bibitem[Chen et~al.(2025{\natexlab{b}})Chen, Shao, Xu, Wang, Gao, Zhang, and
  Luo]{chen2025efficientqat}
Mengzhao Chen, Wenqi Shao, Peng Xu, Jiahao Wang, Peng Gao, Kaipeng Zhang, and
  Ping Luo.
\newblock {EfficientQAT}: Efficient quantization-aware training for large
  language models.
\newblock In \emph{Proceedings of the Annual Meeting of the Association for
  Computational Linguistics (ACL)}, 2025{\natexlab{b}}.

\bibitem[Chen et~al.(2024)Chen, Liu, Wang, Zhang, Liu, Lin, Chen, and
  Zhao]{chen2024agentflan}
Zehui Chen, Kuikun Liu, Qiuchen Wang, Wenwei Zhang, Jiangning Liu, Dahua Lin,
  Kai Chen, and Feng Zhao.
\newblock {Agent-FLAN}: Designing data and methods of effective agent tuning
  for large language models.
\newblock In \emph{Findings of the Association for Computational Linguistics
  (ACL)}, 2024.

\bibitem[Chmiel et~al.(2025)Chmiel, Fishman, Banner, and Soudry]{chmiel2025fp4}
Brian Chmiel, Maxim Fishman, Ron Banner, and Daniel Soudry.
\newblock {FP4} all the way: Fully quantized training of {LLMs}.
\newblock In \emph{Advances in Neural Information Processing Systems
  (NeurIPS)}, 2025.

\bibitem[Choi et~al.(2025)Choi, Lee, Park, Kwon, and Lee]{choi2025falqon}
Kanghyun Choi, Hyeyoon Lee, SunJong Park, Dain Kwon, and Jinho Lee.
\newblock {FALQON}: Accelerating {LoRA} fine-tuning with low-bit floating-point
  arithmetic.
\newblock In \emph{Advances in Neural Information Processing Systems
  (NeurIPS)}, 2025.

\bibitem[Dao and Gu(2024)]{dao2024mamba2}
Tri Dao and Albert Gu.
\newblock Transformers are {SSMs}: Generalized models and efficient algorithms
  through structured state space duality.
\newblock In \emph{International Conference on Machine Learning (ICML)}, 2024.

\bibitem[Dettmers et~al.(2023)Dettmers, Pagnoni, Holtzman, and
  Zettlemoyer]{dettmers2023qlora}
Tim Dettmers, Artidoro Pagnoni, Ari Holtzman, and Luke Zettlemoyer.
\newblock {QLoRA}: Efficient finetuning of quantized {LLMs}.
\newblock In \emph{Advances in Neural Information Processing Systems
  (NeurIPS)}, 2023.

\bibitem[Ding et~al.(2026)Ding, Long, Pu, Zhang, Zhou, Gao,
  et~al.]{ding2025nl2repo}
Jingzhe Ding, Shengda Long, Changxin Pu, Ge~Zhang, Huan Zhou, Hongwan Gao,
  et~al.
\newblock {NL2Repo-Bench}: Towards long-horizon repository generation
  evaluation of coding agents.
\newblock In \emph{International Conference on Machine Learning (ICML)}, volume
  306 of \emph{Proceedings of Machine Learning Research}, pages 25035--25054,
  2026.

\bibitem[Esser et~al.(2020)Esser, McKinstry, Bablani, Appuswamy, and
  Modha]{esser2020lsq}
Steven~K. Esser, Jeffrey~L. McKinstry, Deepika Bablani, Rathinakumar Appuswamy,
  and Dharmendra~S. Modha.
\newblock Learned step size quantization.
\newblock In \emph{International Conference on Learning Representations
  (ICLR)}, 2020.

\bibitem[Frantar et~al.(2023)Frantar, Ashkboos, Hoefler, and
  Alistarh]{frantar2023gptq}
Elias Frantar, Saleh Ashkboos, Torsten Hoefler, and Dan Alistarh.
\newblock {GPTQ}: Accurate post-training quantization for generative
  pre-trained transformers.
\newblock In \emph{International Conference on Learning Representations
  (ICLR)}, 2023.

\bibitem[Guo et~al.(2025)Guo, Li, Dai, Xia, and Benini]{guo2025intlora}
Hang Guo, Yawei Li, Tao Dai, Shu-Tao Xia, and Luca Benini.
\newblock {IntLoRA}: Integral low-rank adaptation of quantized diffusion
  models.
\newblock In \emph{International Conference on Machine Learning (ICML)}, 2025.

\bibitem[Gupta et~al.(2015)Gupta, Agrawal, Gopalakrishnan, and
  Narayanan]{gupta2015limited}
Suyog Gupta, Ankur Agrawal, Kailash Gopalakrishnan, and Pritish Narayanan.
\newblock Deep learning with limited numerical precision.
\newblock In \emph{International Conference on Machine Learning (ICML)}, 2015.

\bibitem[Hu et~al.(2022)Hu, Shen, Wallis, Allen-Zhu, Li, Wang, Wang, and
  Chen]{hu2022lora}
Edward~J. Hu, Yelong Shen, Phillip Wallis, Zeyuan Allen-Zhu, Yuanzhi Li, Shean
  Wang, Lu~Wang, and Weizhu Chen.
\newblock {LoRA}: Low-rank adaptation of large language models.
\newblock In \emph{International Conference on Learning Representations
  (ICLR)}, 2022.

\bibitem[Huang et~al.(2026)Huang, Lee, Tng, and Ge]{datacurve2026deepswe}
Wenqi Huang, Charley Lee, Leonard Tng, and Serena Ge.
\newblock {DeepSWE}: Measuring frontier coding agents on original, long-horizon
  engineering tasks.
\newblock Datacurve blog, \url{https://deepswe.datacurve.ai/blog/deepswe},
  2026.
\newblock Benchmark v1.1, 113 tasks. Paper: arXiv:2607.07946. Code:
  \url{https://github.com/datacurve-ai/deep-swe}.

\bibitem[Ilharco et~al.(2023)Ilharco, Ribeiro, Wortsman, Gururangan, Schmidt,
  Hajishirzi, and Farhadi]{ilharco2023taskarithmetic}
Gabriel Ilharco, Marco~Tulio Ribeiro, Mitchell Wortsman, Suchin Gururangan,
  Ludwig Schmidt, Hannaneh Hajishirzi, and Ali Farhadi.
\newblock Editing models with task arithmetic.
\newblock In \emph{International Conference on Learning Representations
  (ICLR)}, 2023.

\bibitem[Jacob et~al.(2018)Jacob, Kligys, Chen, Zhu, Tang, Howard, Adam, and
  Kalenichenko]{jacob2018quantization}
Benoit Jacob, Skirmantas Kligys, Bo~Chen, Menglong Zhu, Matthew Tang, Andrew
  Howard, Hartwig Adam, and Dmitry Kalenichenko.
\newblock Quantization and training of neural networks for efficient
  integer-arithmetic-only inference.
\newblock In \emph{IEEE/CVF Conference on Computer Vision and Pattern
  Recognition (CVPR)}, 2018.

\bibitem[Jain et~al.(2025)Jain, Han, Gu, Li, Yan, Zhang, Wang, Solar-Lezama,
  Sen, and Stoica]{jain2025livecodebench}
Naman Jain, King Han, Alex Gu, Wen-Ding Li, Fanjia Yan, Tianjun Zhang, Sida
  Wang, Armando Solar-Lezama, Koushik Sen, and Ion Stoica.
\newblock {LiveCodeBench}: Holistic and contamination free evaluation of large
  language models for code.
\newblock In \emph{International Conference on Learning Representations
  (ICLR)}, 2025.

\bibitem[Jeon et~al.(2025)Jeon, Kim, and Kim]{jeon2025l4q}
Hyesung Jeon, Yulhwa Kim, and Jae-Joon Kim.
\newblock {L4Q}: Parameter efficient quantization-aware fine-tuning on large
  language models.
\newblock In \emph{Proceedings of the Annual Meeting of the Association for
  Computational Linguistics (ACL)}, 2025.

\bibitem[Jimenez et~al.(2024)Jimenez, Yang, Wettig, Yao, Pei, Press, and
  Narasimhan]{jimenez2024swebench}
Carlos~E. Jimenez, John Yang, Alexander Wettig, Shunyu Yao, Kexin Pei, Ofir
  Press, and Karthik Narasimhan.
\newblock {SWE-bench}: Can language models resolve real-world {GitHub} issues?
\newblock In \emph{International Conference on Learning Representations
  (ICLR)}, 2024.

\bibitem[Kim et~al.(2023)Kim, Lee, Kim, Park, Yoo, Kwon, and Lee]{kim2023peqa}
Jeonghoon Kim, Jung~Hyun Lee, Sungdong Kim, Joonsuk Park, Kang~Min Yoo, Se~Jung
  Kwon, and Dongsoo Lee.
\newblock Memory-efficient fine-tuning of compressed large language models via
  sub-4-bit integer quantization.
\newblock In \emph{Advances in Neural Information Processing Systems
  (NeurIPS)}, 2023.

\bibitem[Kwon et~al.(2023)Kwon, Li, Zhuang, Sheng, Zheng, Yu, Gonzalez, Zhang,
  and Stoica]{kwon2023vllm}
Woosuk Kwon, Zhuohan Li, Siyuan Zhuang, Ying Sheng, Lianmin Zheng, Cody~Hao Yu,
  Joseph~E. Gonzalez, Hao Zhang, and Ion Stoica.
\newblock Efficient memory management for large language model serving with
  {PagedAttention}.
\newblock In \emph{ACM Symposium on Operating Systems Principles (SOSP)}, 2023.

\bibitem[Lee et~al.(2023)Lee, Chen, Tajwar, Kumar, Yao, Liang, and
  Finn]{lee2023surgical}
Yoonho Lee, Annie~S. Chen, Fahim Tajwar, Ananya Kumar, Huaxiu Yao, Percy Liang,
  and Chelsea Finn.
\newblock Surgical fine-tuning improves adaptation to distribution shifts.
\newblock In \emph{International Conference on Learning Representations
  (ICLR)}, 2023.

\bibitem[Li et~al.(2026)Li, Wang, Huang, and Gotei]{li2026scaleqlora}
Tung-Ling Li, Lee-Chi Wang, Jiale Huang, and Janaki~Ram Gotei.
\newblock {Scale-QLoRA}: Code-invariant adapter merging for native 4-bit
  microscaling {LLMs}.
\newblock arXiv preprint arXiv:2609.04526, 2026.

\bibitem[Li et~al.(2024)Li, Yu, Liang, He, Karampatziakis, Chen, and
  Zhao]{li2024loftq}
Yixiao Li, Yifan Yu, Chen Liang, Pengcheng He, Nikos Karampatziakis, Weizhu
  Chen, and Tuo Zhao.
\newblock {LoftQ}: {LoRA}-fine-tuning-aware quantization for large language
  models.
\newblock In \emph{International Conference on Learning Representations
  (ICLR)}, 2024.

\bibitem[Liu et~al.(2023)Liu, Xie, Li, and Ma]{liu2023samepretrainingloss}
Hong Liu, Sang~Michael Xie, Zhiyuan Li, and Tengyu Ma.
\newblock Same pre-training loss, better downstream: Implicit bias matters for
  language models.
\newblock In \emph{International Conference on Machine Learning (ICML)}, volume
  202 of \emph{Proceedings of Machine Learning Research}, pages 22188--22214,
  2023.

\bibitem[Liu et~al.(2026{\natexlab{a}})Liu, Kong, Dong, Yang, Li, Xie, Gong,
  Shen, Zhao, Tang, Yuan, Niu, Zhang, Lin, Huang, and Wang]{liu2026sad}
Jun Liu, Zhenglun Kong, Peiyan Dong, Changdi Yang, Tianqin Li, Yanyue Xie,
  Yifan Gong, Xuan Shen, Pu~Zhao, Hao Tang, Geng Yuan, Wei Niu, Wenbin Zhang,
  Xue Lin, Dong Huang, and Yanzhi Wang.
\newblock Structured agent distillation for large language model agents.
\newblock In \emph{International Conference on Autonomous Agents and Multiagent
  Systems (AAMAS)}, 2026{\natexlab{a}}.

\bibitem[Liu et~al.(2026{\natexlab{b}})Liu, Sun, He, Zheng, Xu, Zheng, Wang,
  and Chen]{liu2026eksft}
Qi~Liu, Mingdi Sun, Yongyi He, Zhi Zheng, Tong Xu, Yi~Zheng, Zhefeng Wang, and
  Enhong Chen.
\newblock Entropy-{KL} divergence-based token masking: A novel approach for
  selective fine-tuning of large language models.
\newblock arXiv preprint arXiv:2605.29303, 2026{\natexlab{b}}.

\bibitem[Liu et~al.(2024{\natexlab{a}})Liu, Wang, Yin, Molchanov, Wang, Cheng,
  and Chen]{liu2024dora}
Shih-Yang Liu, Chien-Yi Wang, Hongxu Yin, Pavlo Molchanov, Yu-Chiang~Frank
  Wang, Kwang-Ting Cheng, and Min-Hung Chen.
\newblock {DoRA}: Weight-decomposed low-rank adaptation.
\newblock In \emph{International Conference on Machine Learning (ICML)},
  2024{\natexlab{a}}.

\bibitem[Liu et~al.(2024{\natexlab{b}})Liu, Oguz, Zhao, Chang, Stock, Mehdad,
  Shi, Krishnamoorthi, and Chandra]{liu2024llmqat}
Zechun Liu, Barlas Oguz, Changsheng Zhao, Ernie Chang, Pierre Stock, Yashar
  Mehdad, Yangyang Shi, Raghuraman Krishnamoorthi, and Vikas Chandra.
\newblock {LLM-QAT}: Data-free quantization aware training for large language
  models.
\newblock In \emph{Findings of the Association for Computational Linguistics
  (ACL)}, 2024{\natexlab{b}}.

\bibitem[Luo et~al.(2023)Luo, Yang, Meng, Li, Zhou, and
  Zhang]{luo2023forgetting}
Yun Luo, Zhen Yang, Fandong Meng, Yafu Li, Jie Zhou, and Yue Zhang.
\newblock An empirical study of catastrophic forgetting in large language
  models during continual fine-tuning.
\newblock arXiv preprint arXiv:2308.08747, 2023.

\bibitem[{MaxDiffusion contributors}(2026)]{maxdiffusion2026loradelta}
{MaxDiffusion contributors}.
\newblock Keep the full {LoRA} delta when merging into bfloat16 weights.
\newblock GitHub pull request AI-Hypercomputer/maxdiffusion \#513, opened 18
  October, open as of November, 2026.
\newblock URL \url{https://github.com/AI-Hypercomputer/maxdiffusion/pull/513}.
\newblock Accessed 23 November 2026.

\bibitem[Meng et~al.(2022)Meng, Bau, Andonian, and Belinkov]{meng2022rome}
Kevin Meng, David Bau, Alex Andonian, and Yonatan Belinkov.
\newblock Locating and editing factual associations in {GPT}.
\newblock In \emph{Advances in Neural Information Processing Systems
  (NeurIPS)}, 2022.

\bibitem[Meng et~al.(2023)Meng, Sen~Sharma, Andonian, Belinkov, and
  Bau]{meng2023memit}
Kevin Meng, Arnab Sen~Sharma, Alex Andonian, Yonatan Belinkov, and David Bau.
\newblock Mass-editing memory in a transformer.
\newblock In \emph{International Conference on Learning Representations
  (ICLR)}, 2023.

\bibitem[Merrill et~al.(2026)Merrill, Shaw, Carlini, Li, Raj,
  et~al.]{merrill2026terminalbench}
Mike~A. Merrill, Alexander~G. Shaw, Nicholas Carlini, Boxuan Li, Harsh Raj,
  et~al.
\newblock {Terminal-Bench}: Benchmarking agents on hard, realistic tasks in
  command line interfaces.
\newblock In \emph{International Conference on Learning Representations
  (ICLR)}, 2026.
\newblock Version 2.1 of the task set:
  \url{https://tbench.ai/news/terminal-bench-2-1}.

\bibitem[Micikevicius et~al.(2022)Micikevicius, Stosic, Burgess, Cornea, Dubey,
  Grisenthwaite, Ha, Heinecke, Judd, Kamalu, Mellempudi, Oberman, Shoeybi, Siu,
  and Wu]{micikevicius2022fp8}
Paulius Micikevicius, Dusan Stosic, Neil Burgess, Marius Cornea, Pradeep Dubey,
  Richard Grisenthwaite, Sangwon Ha, Alexander Heinecke, Patrick Judd, John
  Kamalu, Naveen Mellempudi, Stuart Oberman, Mohammad Shoeybi, Michael Siu, and
  Hao Wu.
\newblock {FP8} formats for deep learning.
\newblock arXiv preprint arXiv:2209.05433, 2022.

\bibitem[{NVIDIA}(2026)]{nvidia2026nemotron35supervl}
{NVIDIA}.
\newblock {NVIDIA-Nemotron-3.5-Super-VL-120B-A12B} ({BF16} and {NVFP4} model
  cards).
\newblock Hugging Face, 2026.
\newblock URL
  \url{https://huggingface.co/nvidia/NVIDIA-Nemotron-3.5-Super-VL-120B-A12B-NVFP4}.
\newblock The BF16 release is at the same path ending in -BF16. Accessed 23
  November 2026.

\bibitem[{NVIDIA} et~al.(2025{\natexlab{a}}){NVIDIA}, Abecassis, Agrusa, Ahn,
  Alben, et~al.]{nvidia2025nvfp4pretraining}
{NVIDIA}, Felix Abecassis, Anjulie Agrusa, Dong Ahn, Jonah Alben, et~al.
\newblock Pretraining large language models with {NVFP4}.
\newblock arXiv preprint arXiv:2509.25149, 2025{\natexlab{a}}.

\bibitem[{NVIDIA} et~al.(2025{\natexlab{b}}){NVIDIA}, Blakeman, Basant,
  Khattar, et~al.]{nvidia2025nemotronh}
{NVIDIA}, Aaron Blakeman, Aarti Basant, Abhinav Khattar, et~al.
\newblock {Nemotron-H}: A family of accurate and efficient hybrid
  {Mamba-Transformer} models.
\newblock arXiv preprint arXiv:2504.03624, 2025{\natexlab{b}}.

\bibitem[{NVIDIA} et~al.(2026){NVIDIA}, Jhunjhunwala, Srivastava, Gupta,
  et~al.]{nvidia2026nemotron35report}
{NVIDIA}, Aastha Jhunjhunwala, Aayushi~Ashokkumar Srivastava, Abhibha Gupta,
  et~al.
\newblock {Nemotron 3.5 Super} and {Lightning}: Open, efficient
  mixture-of-experts hybrid {Mamba-Transformer} models for agentic {AI}.
\newblock arXiv preprint arXiv:2610.38095, 2026.

\bibitem[{OpenAI}(2024)]{openai2024sweverified}
{OpenAI}.
\newblock Introducing {SWE-bench Verified}.
\newblock Blog post, 13 August, 2024.
\newblock URL \url{https://openai.com/index/introducing-swe-bench-verified/}.
\newblock Accessed 23 November 2026.

\bibitem[Ouyang et~al.(2022)Ouyang, Wu, Jiang, Almeida, Wainwright, Mishkin,
  et~al.]{ouyang2022instructgpt}
Long Ouyang, Jeff Wu, Xu~Jiang, Diogo Almeida, Carroll~L. Wainwright, Pamela
  Mishkin, et~al.
\newblock Training language models to follow instructions with human feedback.
\newblock In \emph{Advances in Neural Information Processing Systems
  (NeurIPS)}, 2022.

\bibitem[Pan et~al.(2025)Pan, Wang, Neubig, Jaitly, Ji, Suhr, and
  Zhang]{pan2025swegym}
Jiayi Pan, Xingyao Wang, Graham Neubig, Navdeep Jaitly, Heng Ji, Alane Suhr,
  and Yizhe Zhang.
\newblock Training software engineering agents and verifiers with {SWE-Gym}.
\newblock In \emph{International Conference on Machine Learning (ICML)}, 2025.

\bibitem[Pyatkin et~al.(2025)Pyatkin, Malik, Graf, Ivison, Huang, Dasigi,
  Lambert, and Hajishirzi]{pyatkin2025ifbench}
Valentina Pyatkin, Saumya Malik, Victoria Graf, Hamish Ivison, Shengyi Huang,
  Pradeep Dasigi, Nathan Lambert, and Hannaneh Hajishirzi.
\newblock Generalizing verifiable instruction following.
\newblock In \emph{Advances in Neural Information Processing Systems (NeurIPS),
  Datasets and Benchmarks Track}, 2025.
\newblock Introduces IFBench.

\bibitem[Qi et~al.(2025)Qi, Liu, Zhou, Pang, Du, Lee, and
  Lin]{qi2025fp16mismatch}
Penghui Qi, Zichen Liu, Xiangxin Zhou, Tianyu Pang, Chao Du, Wee~Sun Lee, and
  Min Lin.
\newblock Defeating the training-inference mismatch via {FP16}.
\newblock arXiv preprint arXiv:2510.26788, 2025.

\bibitem[Qin et~al.(2024)Qin, Ma, Zheng, Li, Zhang, Liu, Luo, Liu, and
  Magno]{qin2024irqlora}
Haotong Qin, Xudong Ma, Xingyu Zheng, Xiaoyang Li, Yang Zhang, Shouda Liu, Jie
  Luo, Xianglong Liu, and Michele Magno.
\newblock Accurate {LoRA}-finetuning quantization of {LLMs} via information
  retention.
\newblock In \emph{International Conference on Machine Learning (ICML)}, 2024.

\bibitem[Rein et~al.(2024)Rein, Hou, Stickland, Petty, Pang, Dirani, Michael,
  and Bowman]{rein2024gpqa}
David Rein, Betty~Li Hou, Asa~Cooper Stickland, Jackson Petty, Richard~Yuanzhe
  Pang, Julien Dirani, Julian Michael, and Samuel~R. Bowman.
\newblock {GPQA}: A graduate-level {Google}-proof {Q\&A} benchmark.
\newblock In \emph{Conference on Language Modeling (COLM)}, 2024.

\bibitem[Ren and Sutherland(2025)]{ren2025learningdynamics}
Yi~Ren and Danica~J. Sutherland.
\newblock Learning dynamics of {LLM} finetuning.
\newblock In \emph{International Conference on Learning Representations
  (ICLR)}, 2025.

\bibitem[Rouhani et~al.(2023)Rouhani, Zhao, More, Hall, Khodamoradi, Deng,
  et~al.]{rouhani2023microscaling}
Bita~Darvish Rouhani, Ritchie Zhao, Ankit More, Mathew Hall, Alireza
  Khodamoradi, Summer Deng, et~al.
\newblock Microscaling data formats for deep learning.
\newblock arXiv preprint arXiv:2310.10537, 2023.

\bibitem[Schulman(2020)]{schulman2020klapprox}
John Schulman.
\newblock Approximating {KL} divergence.
\newblock Blog post, 2020.
\newblock URL \url{http://joschu.net/blog/kl-approx.html}.
\newblock Accessed 23 November 2026.

\bibitem[{SGLang contributors}(2026)]{sglang2026loraunmerged}
{SGLang contributors}.
\newblock [diffusion] fix: keep a {LoRA} unmerged under merge\_mode auto when
  merging would round its update away.
\newblock GitHub pull request sgl-project/sglang \#43385, merged 10 October,
  2026.
\newblock URL \url{https://github.com/sgl-project/sglang/pull/43385}.
\newblock Accessed 23 November 2026.

\bibitem[Shao et~al.(2024)Shao, Wang, Zhu, Xu, Song, Bi, Zhang, Zhang, Li, Wu,
  and Guo]{shao2024deepseekmath}
Zhihong Shao, Peiyi Wang, Qihao Zhu, Runxin Xu, Junxiao Song, Xiao Bi, Haowei
  Zhang, Mingchuan Zhang, Y.~K. Li, Y.~Wu, and Daya Guo.
\newblock {DeepSeekMath}: Pushing the limits of mathematical reasoning in open
  language models.
\newblock arXiv preprint arXiv:2402.03300, 2024.

\bibitem[Shazeer et~al.(2017)Shazeer, Mirhoseini, Maziarz, Davis, Le, Hinton,
  and Dean]{shazeer2017moe}
Noam Shazeer, Azalia Mirhoseini, Krzysztof Maziarz, Andy Davis, Quoc Le,
  Geoffrey Hinton, and Jeff Dean.
\newblock Outrageously large neural networks: The sparsely-gated
  mixture-of-experts layer.
\newblock In \emph{International Conference on Learning Representations
  (ICLR)}, 2017.

\bibitem[Shenfeld et~al.(2026)Shenfeld, Pari, and Agrawal]{shenfeld2025rlrazor}
Idan Shenfeld, Jyothish Pari, and Pulkit Agrawal.
\newblock {RL}'s razor: Why online reinforcement learning forgets less.
\newblock In \emph{International Conference on Learning Representations
  (ICLR)}, 2026.

\bibitem[Singh(2026)]{singh2026vlafp32}
Shubham Singh.
\newblock Full-weight {VLA} fine-tuning without an fp32 master copy silently
  erases most updates.
\newblock arXiv preprint arXiv:2610.31859, 2026.

\bibitem[Tian et~al.(2024)Tian, Gao, Zhang, Chen, Fan, Guo,
  et~al.]{tian2024scicode}
Minyang Tian, Luyu Gao, Shizhuo~Dylan Zhang, Xinan Chen, Cunwei Fan, Xuefei
  Guo, et~al.
\newblock {SciCode}: A research coding benchmark curated by scientists.
\newblock In \emph{Advances in Neural Information Processing Systems (NeurIPS),
  Datasets and Benchmarks Track}, 2024.

\bibitem[White et~al.(2025)White, Dooley, Roberts, Pal, Feuer, Jain,
  Shwartz-Ziv, Jain, Saifullah, Dey, {Shubh-Agrawal}, Sandha, Naidu, Hegde,
  LeCun, Goldstein, Neiswanger, and Goldblum]{white2025livebench}
Colin White, Samuel Dooley, Manley Roberts, Arka Pal, Benjamin Feuer,
  Siddhartha Jain, Ravid Shwartz-Ziv, Neel Jain, Khalid Saifullah, Sreemanti
  Dey, {Shubh-Agrawal}, Sandeep~Singh Sandha, Siddartha Naidu, Chinmay Hegde,
  Yann LeCun, Tom Goldstein, Willie Neiswanger, and Micah Goldblum.
\newblock {LiveBench}: A challenging, contamination-limited {LLM} benchmark.
\newblock In \emph{International Conference on Learning Representations
  (ICLR)}, 2025.
\newblock Leaderboard (version 2.0): \url{https://livebench.ai}.

\bibitem[Wu and Xie(2024)]{wu2024vstar}
Penghao Wu and Saining Xie.
\newblock {V*}: Guided visual search as a core mechanism in multimodal {LLMs}.
\newblock In \emph{Proceedings of the IEEE/CVF Conference on Computer Vision
  and Pattern Recognition (CVPR)}, 2024.

\bibitem[Xin et~al.(2026)Xin, Priyadarshi, Xin, Kartal, Vavre,
  et~al.]{xin2026qad}
Meng Xin, Sweta Priyadarshi, Jingyu Xin, Bilal Kartal, Aditya Vavre, et~al.
\newblock Quantization-aware distillation for {NVFP4} inference accuracy
  recovery.
\newblock arXiv preprint arXiv:2601.20088, 2026.
\newblock NVIDIA technical report.

\bibitem[Xu et~al.(2024)Xu, Xie, Gu, Chen, Chang, Zhang, Chen, Zhang, and
  Tian]{xu2024qalora}
Yuhui Xu, Lingxi Xie, Xiaotao Gu, Xin Chen, Heng Chang, Hengheng Zhang, Zhengsu
  Chen, Xiaopeng Zhang, and Qi~Tian.
\newblock {QA-LoRA}: Quantization-aware low-rank adaptation of large language
  models.
\newblock In \emph{International Conference on Learning Representations
  (ICLR)}, 2024.

\bibitem[Yadav et~al.(2023)Yadav, Tam, Choshen, Raffel, and
  Bansal]{yadav2023ties}
Prateek Yadav, Derek Tam, Leshem Choshen, Colin Raffel, and Mohit Bansal.
\newblock {TIES-Merging}: Resolving interference when merging models.
\newblock In \emph{Advances in Neural Information Processing Systems
  (NeurIPS)}, 2023.

\bibitem[Yang et~al.(2025)Yang, Lieret, Jimenez, Wettig, Khandpur, Zhang, Hui,
  Press, Schmidt, and Yang]{yang2025swesmith}
John Yang, Kilian Lieret, Carlos~E. Jimenez, Alexander Wettig, Kabir Khandpur,
  Yanzhe Zhang, Binyuan Hui, Ofir Press, Ludwig Schmidt, and Diyi Yang.
\newblock {SWE-smith}: Scaling data for software engineering agents.
\newblock In \emph{Advances in Neural Information Processing Systems (NeurIPS),
  Datasets and Benchmarks Track}, 2025.
\newblock Introduces SWE-bench Multilingual.

\bibitem[Yao et~al.(2025)Yao, Liu, Zhang, Dong, Shang, and
  Gao]{yao2025offpolicy}
Feng Yao, Liyuan Liu, Dinghuai Zhang, Chengyu Dong, Jingbo Shang, and Jianfeng
  Gao.
\newblock Your efficient {RL} framework secretly brings you off-policy {RL}
  training.
\newblock Blog post, 5 August, 2025.
\newblock \mbox{\url{https://fengyao.notion.site/off-policy-rl}}. Code:
  \url{https://github.com/yaof20/Flash-RL}. Accessed 23 November 2026.

\bibitem[Yu et~al.(2024)Yu, Yu, Yu, Huang, and Li]{yu2024dare}
Le~Yu, Bowen Yu, Haiyang Yu, Fei Huang, and Yongbin Li.
\newblock Language models are super mario: Absorbing abilities from homologous
  models as a free lunch.
\newblock In \emph{International Conference on Machine Learning (ICML)}, 2024.

\bibitem[Zamirai et~al.(2020)Zamirai, Zhang, Aberger, and
  De~Sa]{zamirai2020bf16}
Pedram Zamirai, Jian Zhang, Christopher~R. Aberger, and Christopher De~Sa.
\newblock Revisiting {BFloat16} training.
\newblock arXiv preprint arXiv:2010.06192, 2020.

\bibitem[Zeng et~al.(2024)Zeng, Liu, Lu, Wang, Liu, Dong, and
  Tang]{zeng2024agenttuning}
Aohan Zeng, Mingdao Liu, Rui Lu, Bowen Wang, Xiao Liu, Yuxiao Dong, and Jie
  Tang.
\newblock {AgentTuning}: Enabling generalized agent abilities for {LLMs}.
\newblock In \emph{Findings of the Association for Computational Linguistics
  (ACL)}, 2024.

\bibitem[Zhang et~al.(2026)Zhang, Makhija, Arivazhagan, Kumar, and
  Gangadharaiah]{zhang2026actobs}
Juzheng Zhang, Disha Makhija, Manoj~Ghuhan Arivazhagan, Vinayshekhar~Bannihatti
  Kumar, and Rashmi Gangadharaiah.
\newblock Don't mask the environment: Observation supervision changes how
  agents explore under {RL}.
\newblock arXiv preprint arXiv:2609.20715, 2026.

\bibitem[Zhang et~al.(2025)Zhang, Wang, Li, Wu, Tang, Liu, He, Yin, and
  Wang]{zhang2025unlearningquant}
Zhiwei Zhang, Fali Wang, Xiaomin Li, Zongyu Wu, Xianfeng Tang, Hui Liu, Qi~He,
  Wenpeng Yin, and Suhang Wang.
\newblock Catastrophic failure of {LLM} unlearning via quantization.
\newblock In \emph{International Conference on Learning Representations
  (ICLR)}, 2025.

\end{thebibliography}

\clearpage
\appendix
\section*{Appendix}
\section{Training configuration}
\label{app:config}

\Cref{tab:hparams} lists the settings of the two released runs.
Both runs see the same rows in the same order.
They share the objective, the LoRA rank, the optimizer and the schedule.
The only differences come from how NVIDIA stores each base model.
Two module groups are adapted in both runs: the attention projections ($q,k,v,o$ in the 8 attention layers) and the Mamba input
projections (\texttt{in\_proj} in the 40 Mamba layers).
These weights are BF16 in the BF16 base and FP8 in the NVFP4 base.
So the NVFP4 run trains them with a four times larger learning rate, which moves them as far as the BF16 weights move
(\cref{sec:ratetarget}).
The third group is different in the two runs.
The BF16 run adapts the up and down projections of the shared experts in the 40 MoE layers.
In the NVFP4 base these weights are stored in 4 bits and do not move at an update of our size (\cref{app:nvfp4}).
So the NVFP4 run adapts the MoE latent projections (fc1 and fc2) instead, which NVIDIA stores in BF16.
Each run emulates the serving numerics of its own format.

\begin{table}[h]
\centering
\caption{\textbf{Training configuration} of the two released 120B models. Rows that span both columns are identical in the two
runs. Numbers in parentheses are counts of adapted weight matrices, and for trainable parameters the share of all model
parameters.}
\label{tab:hparams}
\footnotesize
\begin{tabular}{>{\raggedright\arraybackslash}p{3.6cm}>{\raggedright\arraybackslash}p{5.6cm}>{\raggedright\arraybackslash}p{5.6cm}}
\toprule
 & \smaug{} (BF16) & \smaugfp{} \\
\midrule
adapted modules & $q,k,v,o$ (32), \texttt{in\_proj} (40), shared up/down (80) & $q,k,v,o$ (32), \texttt{in\_proj} (40), latent fc1/fc2 (80) \\
storage of adapted modules & BF16 (152) & FP8 (72), BF16 (80) \\
trainable parameters & 59.7M (0.049\%) & 48.5M (0.040\%) \\
\midrule
LoRA & \multicolumn{2}{p{11.2cm}}{rank 32, $\alpha = 64$ (scale $s = 2$), dropout 0} \\
objective & \multicolumn{2}{p{11.2cm}}{likelihood on action tokens only; sequence-level $\kthree$ drift penalty on thinking, $\beta = 100$} \\
\midrule
optimizer & \multicolumn{2}{p{11.2cm}}{AdamW, weight decay 0, gradient clipping at 1.0} \\
peak learning rate & $6\times10^{-7}$ & $4.5\times10^{-7}$; FP8 modules $\times4$ ($\gamma_{\mathrm{fp8}}$) \\
schedule & \multicolumn{2}{p{11.2cm}}{linear warmup over 16 steps, then cosine decay to $0.3\times$ the peak at step 158} \\
steps $\times$ rows & \multicolumn{2}{p{11.2cm}}{158 steps $\times$ 24 rows, one pass over the 3{,}767 rows in the same order (25 rows seen twice); rows up to 65{,}536 tokens, not packed} \\
\midrule
serving numerics in training & FP8 $Q,K,V$, Mamba $\Delta t$ clamp removed & the same, plus FP8 W8A8, NVFP4 W4A4 experts, MoE combine \\
released checkpoint & step 100 ($\rho = 3.1\times10^{-5}$) & step 100 ($\rho = 3.5\times10^{-5}$) \\
\bottomrule
\end{tabular}
\end{table}

\section{Merge validation}
\label{app:validation}

\paragraph{Emulated serving operations.}
The training forward pass reproduces the operations of the serving engine that touch the adapted weights.
For the BF16 run these are two.
The engine rounds $K$ and $V$ to FP8 E4M3 when it writes them to the KV cache, and $Q$ at the input of attention. We apply the same
rounding in training, with a straight-through gradient.
The engine applies no lower bound to the Mamba time step, where the training library clamps it at $10^{-3}$. We remove the clamp.
For the NVFP4 run we add the FP8 linears with their served activation quantization, the NVFP4 shared experts with 4-bit weights and
activations, the routed-expert kernel with its BF16 rounding order, and the approximate reciprocals the engine uses to compute scales.
Three known deviations are not emulated: the FP32 residual stream, the tile-wise quantization of the softmax, and the FP8 rounding of
the attention output before the output projection. None of them depends on the adapter.
So the exactness claims of this appendix are about weights. Activations are matched operation by operation, not bit for bit.

\paragraph{Export checks.}
To verify that the trained model is the shipped model, we recompute every adapted value of the released checkpoints from the base
checkpoint and the saved adapter, independently of the trainer and of the merge script, and compare it with the released files.
All 5.09 billion adapted elements of the BF16 release and all 3.66 billion of the NVFP4 release match, with no mismatches.
Every tensor we did not adapt, and the configuration, tokenizer and chat template files, are byte for byte identical to the base.
The number of values that changed matches the count the trainer logged at every checkpoint, which ties the weights the loss was computed
on to the weights in the files.
The trainer forms $W + sBA$ in FP32 and rounds it onto the storage grid with the vendor's frozen scales, and the export does the same,
so we rely on this count for their agreement rather than on an element by element comparison.

\section{Served change per checkpoint}
\label{app:rho}

\Cref{tab:rho} lists $\rho$ for the checkpoints of both runs, and for the BF16 run the Terminal-Bench 2.1 score of each merged
checkpoint.
$\rho$ grows steadily with training in both runs, so choosing a checkpoint by $\rho$ means choosing how far the shipped model moves.
During development we merged the BF16 checkpoints at steps 75, 100, 125 and 158 and scored each.
Step 100 solved the most tasks. The smaller update at step 75 had not changed the model enough, and the larger updates at steps 125 and
158 began to cost performance.
This is the evidence behind the range of $\rho$, $2.5$ to $3.5\times10^{-5}$, that we use as a target (\cref{sec:ratetarget,sec:target}).
The NVFP4 checkpoint was then chosen by $\rho$ alone, and its step 100 sits in the same range.

\begin{table}[h]
\centering
\caption{\textbf{Served change per checkpoint.} $\rho$: change of the served weights relative to the base weights (\cref{sec:ratetarget}).
TB 2.1: Terminal-Bench 2.1 tasks solved on a common subset of 67 tasks (step 100: mean of four runs; other steps: one run). Bold:
released checkpoints. The last NVFP4 row is step 155, the last step at which the trainer logged the served change.}
\label{tab:rho}
\small
\begin{tabular}{rcc|rc}
\toprule
\multicolumn{3}{c|}{\smaug{} (BF16)} & \multicolumn{2}{c}{\smaugfp{}} \\
step & $\rho$ & TB 2.1 (of 67) & step & $\rho$ \\
\midrule
75  & $2.24\times10^{-5}$ & 46 (69\%) & 80  & $2.63\times10^{-5}$ \\
\textbf{100} & $\mathbf{3.09\times10^{-5}}$ & \textbf{52.5} (78\%) & \textbf{100} & $\mathbf{3.46\times10^{-5}}$ \\
125 & $3.75\times10^{-5}$ & 51 (76\%) & 110 & $3.82\times10^{-5}$ \\
158 & $4.38\times10^{-5}$ & 49 (73\%) & 130 & $4.44\times10^{-5}$ \\
    &                     &           & 155 & $5.06\times10^{-5}$ \\
\bottomrule
\end{tabular}
\end{table}

\section{Why NVFP4 tensors do not move at our update size}
\label{app:nvfp4}
An NVFP4 weight is stored as a 4-bit code from the set $\{0, 0.5, 1, 1.5, 2, 3, 4, 6\}$.
The code is multiplied by an FP8 scale shared by a block of 16 elements and by one FP32 scale for the whole tensor.
So neighboring grid values are $1/12$ to $1/3$ of the block maximum apart.
For comparison, the relative spacing is roughly $6$ to $12\%$ for FP8 E4M3 and $0.4$ to $0.8\%$ for BF16.
A weight changes its stored code only when the update carries it past the midpoint between two grid values.
Our update moves weights by about as much as in the BF16 run.
For the shared experts, the $99.9$th percentile of $|D_{ij}|$ is only $0.007$ of half a grid step.
So the codes do not change at our update size, and about $0.1\%$ of them would flip even with a learning rate multiplier of about 140.

\section{Compute}
The BF16 run took 2.94 hours on 8 B300 GPUs, including startup and checkpoint saves.
A step took 65 seconds, and peak memory was 117 GB per GPU.
The NVFP4 run took 5.1 hours on 4 B300 GPUs, with 114 seconds per step and 136 GB peak memory.
The plain LoRA run of \cref{sec:history} took 1.7 hours for its 100 steps on 8 B200 GPUs, at 59 seconds per step.

\section{Trajectory judge}
\label{app:judge}
This appendix gives the method and the full results behind \cref{sec:trajectories} and the lower block of \cref{tab:behavior}.

\paragraph{Per-step statistics.}
We first looked at single model calls.
Over all base and \smaug{} runs on the 89 tasks (three and four runs), \smaug{} spends about as many reasoning tokens per call as the
base. The medians over task-runs are 2{,}055 and 2{,}144 tokens.
Its actions have the same length, 401 against 390 tokens by the same statistic.
So the difference between the models is not visible in a single call.
This is why we judged whole trajectories.

\paragraph{Taxonomy.}
We use the failure taxonomy of the Terminal-Bench 2.0 paper~\citep{merrill2026terminalbench}. Its authors derived it from an
earlier taxonomy of multi-agent failures, dropped or merged the modes that do not apply to a single agent in a terminal, and grouped the nine that remain in
three classes: execution, coherence and verification.
Execution modes: the agent disobeys the specification, repeats a step, or does not stop when it should. The last covers continuing after
success or futility and declaring completion before the task is done.
Coherence modes: the agent's commands contradict its stated analysis, it loses earlier context, or it drifts to another goal.
Verification modes: the agent stops before an explicit requirement is met, it does not verify (for example it displays a file instead of
running it), or it verifies weakly, with checks that miss a stated requirement.
A trajectory can carry several modes.
A class row (``any'') is the share of trajectories that carry at least one of the three modes of that class. A trajectory is counted
once even if it carries two or three of them, so a class row is at most the sum of its three modes and usually less.
``Premature completion'' is the union of unaware of termination and premature termination: the agent declared the task complete while
an explicit requirement was visibly unmet.
``No failure mode'' means none of the nine.

\paragraph{Trajectories judged.}
We judge three runs per model under the same harness (\cref{sec:setup}): the base's three runs, three of the plain adapter's four
and three of the release's four.
The judged set is every task that at least one of these nine runs failed. That is 59 tasks.
The judge declined one task for every model, which leaves 58 tasks.
On these tasks we have 174 base, 173 plain LoRA and 174 release trajectories. One plain LoRA trajectory was declined.
Every trajectory on these tasks is judged, whether it passed or failed. 79, 81 and 91 of them passed.

\paragraph{Judge.}
Claude Opus 5.5, run as an agent, serves as the judge. The rubric text is the Terminal-Bench paper's own, with two additions by us:
a decision procedure for task derailment, which the paper does not print, and the abbreviated steps of weak verification. The paper
applies one rubric per mode. We decide all nine modes in one call per trajectory, with written judging conventions
that say what counts as a repetition, when a verification is weak, and how to treat an episode that ran out of time, and with worked
examples.
The judge is given the task description, the agent's own harness instructions and a condensed transcript.
In the transcript every action turn is numbered. The turns around the first completion claim, the writing of the deliverable and the
end of the episode are kept nearly in full. Other turns are trimmed to their head and tail, and long waits and repeated polling are
elided.
The judge assigns the nine modes from the transcript with the hidden reasoning removed. What it sees is the agent's visible analysis
and plan, its commands and the terminal output. This follows the Terminal-Bench paper, which defines the taxonomy on the agent's stated
analysis and its observable actions.
For failed trajectories the judge is told which tests failed. For passed trajectories the outcome is withheld, so the judge cannot call
a trajectory clean because it knows the task passed.
Every label cites the turns it rests on. A validator checks that the cited turns exist in the transcript.

\paragraph{Calibration.}
The Terminal-Bench authors validated their judge against human annotators. We did not label trajectories by hand. Instead we
checked the judge's consistency in two ways.
First, twenty failed trajectories were judged again with the outcome withheld. The judge agreed with itself on 93\% of the mode
decisions.
Second, ten trajectories on tasks that neither the base nor \smaug{} ever fails were judged as well, to check that the judge does not read success as the absence of faults. Four carry no mode, four a
verification mode and two a step repetition.

\begin{table}[tp]
\centering
\caption{\textbf{All judged trajectories} on the 58 judged tasks: share of each model's trajectories (174, 173 and 174) that carry each
mode, with the range of the three single-run shares in parentheses. Lower is better for every failure mode and class, and higher is
better for the last row. A trajectory can carry several modes or none, so the columns do not sum to a constant. A class row (``any'') is the share of
trajectories with at least one mode of that class, each trajectory counted once, so it is less than the sum of the three modes below it
whenever a trajectory carries more than one. All nine modes and the three
classes are shown; within each class the modes are ordered by the release's improvement over the base. In every row other than step repetition, the execution class that contains it, and no failure
mode, the range of the release overlaps the range of the base, so the differences in these rows are within the run-to-run variation of a
single model.}
\label{tab:app_modes}
\small
\begin{tabular}{lccc}
\toprule
 & base & plain LoRA & release \\
\midrule
\textbf{coherence class, any (lower is better)}                         &  27.6 (25.9--31.0) &  28.9 (22.8--34.5) &  25.9 (24.1--27.6) \\
\quad context loss                                                      &  21.3 (19.0--24.1) &  20.2 (15.8--24.1) &  19.0 (17.2--20.7) \\
\quad reasoning-action mismatch                                       &  14.4 (13.8--15.5) &   13.9 (8.8--22.4) &  13.2 (10.3--15.5) \\
\quad task derailment                                                   &     0.6 (0.0--1.7) &     0.0 (0.0--0.0) &     0.0 (0.0--0.0) \\
\midrule
\textbf{execution class, any (lower is better)}                         &  51.1 (50.0--53.4) &  45.7 (43.1--50.0) &  44.8 (41.4--48.3) \\
\quad step repetition                                                   &  45.4 (41.4--50.0) &  37.6 (31.6--43.1) &  36.8 (32.8--39.7) \\
\quad unaware of termination                                            &  19.0 (13.8--24.1) &  20.8 (17.2--27.6) &  20.7 (19.0--24.1) \\
\quad disobeys the specification                                        &   13.2 (8.6--19.0) &  15.0 (12.1--19.0) &  15.5 (15.5--15.5) \\
\midrule
\textbf{verification class, any (lower is better)}                      &  57.5 (51.7--62.1) &  61.3 (59.6--63.8) &  59.8 (56.9--63.8) \\
\quad no or incorrect verification                                      &  29.9 (29.3--31.0) &  32.9 (31.0--36.8) &  27.6 (25.9--31.0) \\
\quad premature termination                                             &  14.4 (12.1--19.0) &  17.3 (15.5--20.7) &  15.5 (13.8--17.2) \\
\quad weak verification                                                 &  27.0 (20.7--32.8) &  27.7 (22.8--31.0) &  31.0 (25.9--36.2) \\
\midrule
\textbf{no failure mode (higher is better)}                             &  15.5 (13.8--17.2) &  17.9 (17.2--19.0) &  23.0 (20.7--25.9) \\
\bottomrule
\end{tabular}
\end{table}

\paragraph{Reading the table.}
The three models share one failure profile. Verification modes occur on more than half of the trajectories, execution modes on about
half, and coherence modes on about a quarter. The update changes this profile at one point, step repetition and the execution
class that contains it, and raises the share of trajectories with no mode at all.
Not every difference in this table is a real change.
Each model was run three times, and the share of a mode moves from run to run.
So we compare the range over the three runs, not the single average.
Three rows stand out.
In step repetition, and in the execution class that contains it, the worst of the three release runs is still better than the best of
the three base runs.
In the share of trajectories with no failure mode, the lowest release run is above the highest base run.
In these three rows the two ranges do not touch.
These are the changes the update made.
In every other row the two ranges overlap.
The release may look a little better or a little worse on average, but the base's own runs spread at least as far.
We read these rows as unchanged.
The verification class is the row most likely to be misread.
The release averages 59.8 against 57.5 for the base, and weak verification 31.0 against 27.0.
But the base's three runs span 51.7 to 62.1 on the class row and 20.7 to 32.8 on weak verification, so the release sits inside the
base's own spread.
Within the class, the release shows slightly less ``no verification'' and slightly more ``weak verification''.
A few trajectories moved from not checking at all to checking weakly.
The release verifies as often as the base, no better and no worse.

\paragraph{By outcome.}
The failed trajectories of the three models look alike.
About half repeat steps (54, 52 and 52\%). A third to a half carry a coherence mode (36, 43 and 45\%). A third declare completion with
a requirement visibly unmet (35, 36 and 37\%).
The passed trajectories differ from the base's.
When the base passes, it still repeats steps in 35\% of its runs and carries a coherence mode in 18\%. The plain adapter does so in
21\% and 12\%, \smaug{} in 23\% and 9\%.
The 26 tasks that the base failed and \smaug{} passed at least once tell the same story from the other side.
On these tasks, the base's failed runs carry a coherence mode in 24\% and a premature completion in 24\%. \smaug{}'s passed runs carry
them in 2\% and 9\%. \smaug{}'s own failed runs on the same tasks carry them in 38\% and 24\%.
So the difference lies between passing and failing runs, not between the models.

\end{document}